\def\year{2020}\relax
\documentclass[letterpaper]{article} 
\usepackage{aaai20}  
\usepackage{times}  
\usepackage{helvet} 
\usepackage{courier}  
\usepackage[hyphens]{url}  
\usepackage{graphicx} 
\usepackage{graphicx}  
\usepackage{times}
\usepackage{epsfig}
\usepackage{amsmath}
\usepackage{amssymb}
\usepackage{lipsum}
\usepackage{etoolbox}
\usepackage{subfig}
\usepackage[ruled]{algorithm2e}
\usepackage{array}
\usepackage{comment}
\usepackage{booktabs}
\newcolumntype{P}[1]{>{\centering\arraybackslash}p{#1}}
\newcolumntype{M}[1]{>{\centering\arraybackslash}m{#1}}
\usepackage{bbm}

\DeclareMathOperator*{\argmin}{arg\,min}
\DeclareMathOperator*{\argmax}{arg\,max}

\newcommand{\eg}{\textit{e.g.} }
\newcommand{\ie}{\textit{i.e.} }
\newcommand{\etc}{etc. }
\newcommand{\etal}{\textit{et al.} }

\newcommand{\figLabel}{Figure\xspace}
\newcommand{\eqLabel}[1]{Eq (#1)}
\newcommand{\secLabel}{Section\xspace}

\newcommand{\mysection}[1]{\vspace{2pt}\noindent\textbf{#1.}}

\newcommand{\supp}{\textbf{supplement}\xspace}
\newcommand{\specialcell}[2][c]{%
  \begin{tabular}[#1]{@{}c@{}}#2\end{tabular}}

\title{SADA: Semantic Adversarial Diagnostic Attacks for Autonomous Applications}

\author{Abdullah Hamdi, Matthias M\"uller, Bernard Ghanem\\ \\
King Abdullah University of Science and Technology (KAUST), Thuwal, Saudi Arabia\\ \\
{\tt\small \{abdullah.hamdi, matthias.mueller.2, bernard.ghanem\} @kaust.edu.sa}
}

\begin{document}

\maketitle

\begin{abstract}
One major factor impeding more widespread adoption of deep neural networks (DNNs) is their lack of robustness, which is essential for safety-critical applications such as autonomous driving. This has motivated much recent work on adversarial attacks for DNNs, which mostly focus on pixel-level perturbations void of semantic meaning. In contrast, we present a general framework for adversarial attacks on trained agents, which covers semantic perturbations to the environment of the agent performing the task as well as pixel-level attacks. To do this, we re-frame the adversarial attack problem as learning a distribution of parameters that always fools the agent. In the semantic case, our proposed adversary (denoted as BBGAN) is trained to sample parameters that describe the environment with which the black-box agent interacts, such that the agent performs its dedicated task poorly in this environment. We apply BBGAN on three different tasks, primarily targeting aspects of autonomous navigation: object detection, self-driving, and autonomous UAV racing. On these tasks, BBGAN can generate failure cases that consistently fool a trained agent.
\end{abstract}

\section{Introduction} \label{intro}
As a result of recent advances in machine learning and computer vision, deep neural networks (DNNs) are now interleaved with many aspects of our daily lives. DNNs suggest news articles to read and movies to watch, automatically edit our photos and videos, and translate between hundreds of languages. They are also bound to disrupt transportation with autonomous driving slowly becoming a reality. While there are already impressive demos and some successful deployments, safety concerns for boundary conditions persist. While current models work very well on average, they struggle with robustness in certain cases. 
\begin{figure}[t]
\centering
   \includegraphics[width=\columnwidth]{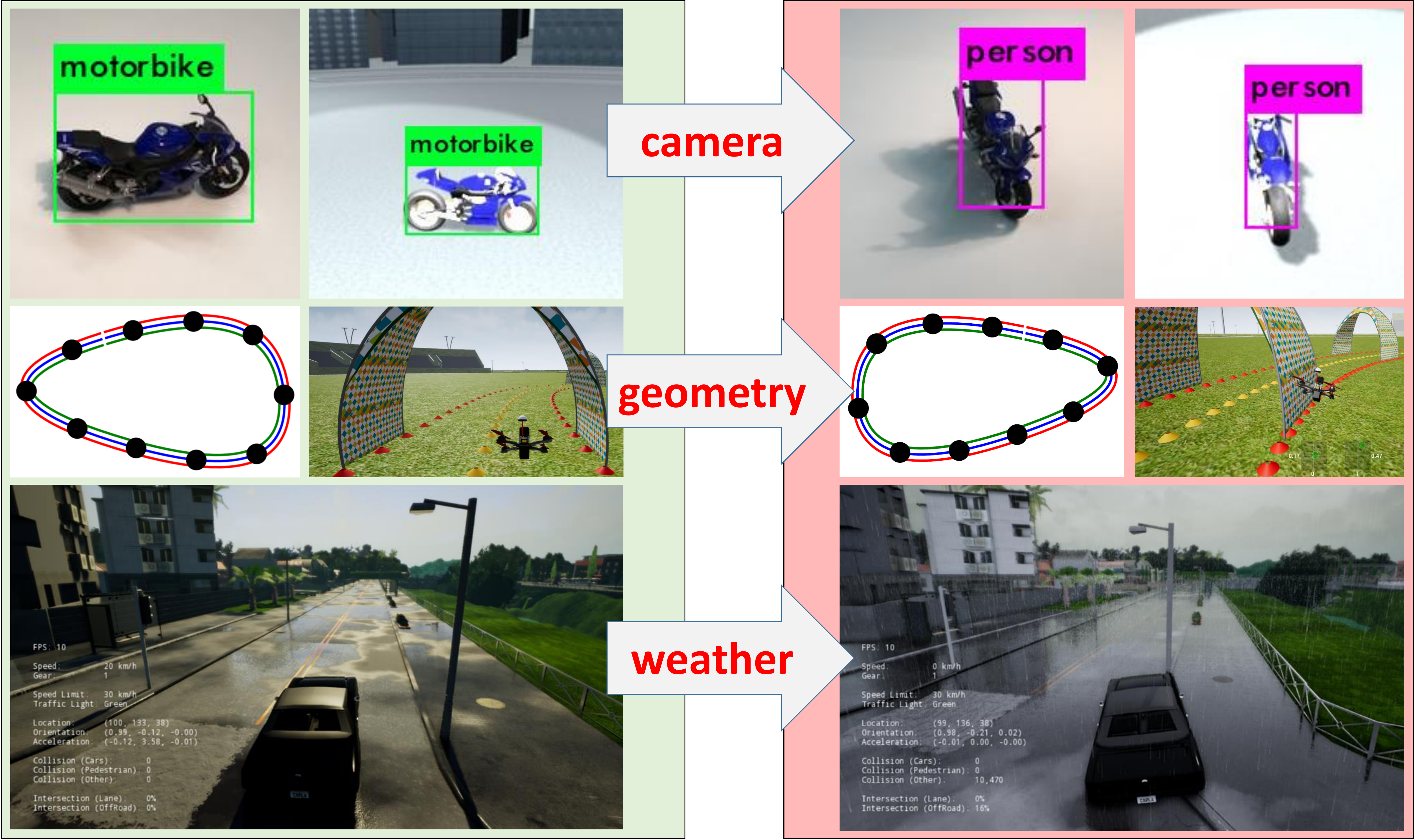}
   \caption{\small \textbf{Semantic Adversarial Diagnostic Attacks}. Neural networks can perform very well on average for a host of tasks; however, they do perform poorly or downright fail when encountering some environments. To diagnose why they fail and how they can be improved, we seek to learn the underlying distribution of semantic parameters, which generate environments that pose difficulty to these networks when applied to three safety critical tasks: 
   object detection, self-driving cars, and autonomous UAV racing.}
   \label{fig:intro_fig}
\end{figure}
Recent work in the adversarial attack literature shows how sensitive DNNs are to input noise. These attacks usually utilize the information about the network structure to perform gradient updates in order to derive targeted perturbations (coined white-box attacks).
These perturbations are injected into the input image at the pixel-level, so as to either confuse the network or enforce a specific behavior \cite{first-attack,fast-sign,carlini,projected-gradient}.

In practice, such pixel attacks are less likely to naturally occur than semantic attacks which include changes in camera viewpoint, lighting conditions, street layouts, etc. The literature on semantic attacks is much sparser, since they are much more subtle and  difficult to analyze \cite{strike,advpose}. Yet, this type of attack is critical to understand/diagnose failure cases that might occur in the real-world. While it is very difficult to investigate semantic attacks on real data, we can leverage simulation as a proxy that can unearth useful insights transferable to the real-world. \figLabel \ref{fig:intro_fig} shows an example of an object misclassified by the YOLOV3 detector \cite{yolo3} applied to a rendered image from a virtual environment, an autonomous UAV racing \cite{oil} failure case in a recently developed general purpose simulator (Sim4CV \cite{sim4cv}), and an autonomous driving failure case in a popular driving simulator (CARLA \cite{carla}). These failures arise from adversarial attacks on the semantic parameters of the environment.

In this work, we consider environments that are adequately photo-realistic and parameterized by a compact set of variables that have direct semantic meaning (\eg camera viewpoint, lighting/weather conditions, road layout, \etc). Since the generation process of these environments from their parameters is quite complicated and in general non-differentiable, we treat it as a \emph{black-box} function that can be queried but not back-propagated through. We seek to learn an adversary that can produce fooling parameters to construct an environment where the agent (which is also a black-box) fails in its task. Unlike most adversarial attacks that generate sparse instances of failure, our proposed adversary provides a more comprehensive view on how an agent can fail; we learn the distribution of fooling parameters for a particular agent and task and then sample from it. Since Generative Adversarial Networks (GANs \cite{GAN,WGAN}) have emerged as a promising family of unsupervised learning techniques that can model high-dimensional distributions, we model our adversary as a GAN, denoted as black-box GAN (BBGAN). 

\mysection{Contributions}
(1) We formalize adversarial attacks in a more general setup to include both semantic and conventional pixel attacks. (2) We propose BBGAN in order to learn the underlying distribution of semantic adversarial attacks and show promising results on three different safety-critical applications used in autonomous navigation.

\section{Related Work}
\subsection{Pixel-level Adversarial Attacks}
Szegedy \etal formulate attacking neural networks as an optimization problem \cite{first-attack}. Their method produces a minimal perturbation of the image pixels that fools a trained classifier (incorrect predictions). Several works followed the same approach but with different formulations, such as Fast Gradient Sign Method (FGSM) \cite{fast-sign} and Projected Gradient Descent \cite{projected-gradient}. A comprehensive study of different ways to fool networks with minimal pixel perturbation can be found in the paper \cite{carlini}. Most efforts use the gradients of the function to optimize the input, which might be difficult to obtain in some cases \cite{advpose}. However, all of these methods are limited to pixel perturbations to fool trained agents, while we consider more general cases of attacks, \eg changes in camera viewpoint to fool a detector or change in weather conditions to fool a self-driving car. Furthermore, we are interested in the distribution of the semantic parameters that fool the agent, more so than individual fooling examples.

\subsection{Semantic Attacks beyond Pixels}
Beyond pixel perturbations,
several recent works perform attacks on the object/camera pose to fool a classifier \cite{strike,advpose,semanticrobustness}. Other works proposed attacks on 3D point clouds using Point-Net \cite{advpoint}, and on 3D meshes using differentiable functions that describe the scene \cite{advmesh}. 
Inspired by these excellent works, we extend semantic attacks by using readily available virtual environments with plausible 3D setups to systematically test trained agents. In fact, our formulation includes attacks not only on static agents like object detectors, but also agents that interact with dynamic environments, such as self-driving agents. To the best of our knowledge, this is the first work to introduce adversarial attacks in CARLA\cite{carla}, a standard autonomous navigation benchmark. 

\begin{figure}[t!]
   \centering
   \includegraphics[page=1 ,width=\columnwidth]{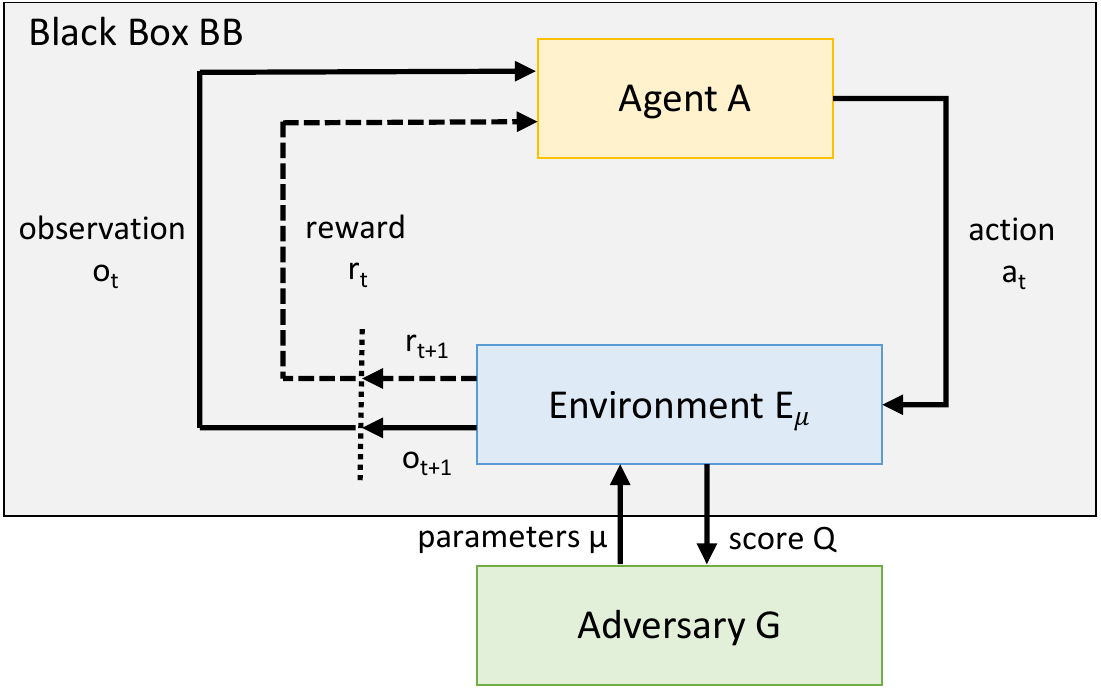}
   \caption{\small \textbf{Generic Adversarial Attacks on Agents}. $\mathbf{E}_{\boldsymbol{\mu}}$ is a parametric environment with which an agent $\mathbf{A}$ interacts. The agent receives an observation $\mathbf{o_{t}}$ from the environment and produces an action $\mathbf{a_{t}}$. The environment scores the agent and updates its state until the episode finishes. A final score $\mathit{Q}(\mathbf{A},\mathbf{E}_{\boldsymbol{\mu}})$ is given to the adversary $\mathbf{G}$, which in turn updates itself to propose more adversarial  parameters $\boldsymbol{\mu}$ for the next episode.}
   \label{fig:setup}
\end{figure}

\subsection{Adversarial Attacks and Reinforcement Learning}
Our generic formulation of adversarial attacks is naturally inspired by RL, in which agents can choose from multiple actions and receive partial rewards as they proceed in their task \cite{reward}. In RL, the agent is subject to training in order to achieve a goal in the environment; the environment can be dynamic to train a more robust agent \cite{robust-rl,robust-adversarial-rl,rlgoal}. However, in adversarial attacks, the agent is usually fixed and the adversary is the subject of the optimization in order to fool the agent. We formulate adversarial attacks in a general setup where the environment rewards an agent for some task. An adversary outside the environment is tasked to fool the agent by modifying the environment and receiving a score after each episode.

\section{Methodology} \label{methodology}
\subsection{Generalizing Adversarial Attacks} \label{sec: attacks}
\mysection{Extending attacks to general agents} 
In this work, we generalize the adversarial attack setup beyond pixel perturbations. Our more general setup (refer to \figLabel{ \ref{fig:setup}}) includes semantic attacks, \eg perturbing the camera pose or lighting conditions of the environment that generates observations (\eg pixels in 2D images). An environment $\mathbf{E}_{\boldsymbol{\mu}}$ is parametrized by $\boldsymbol{\mu} \in [\boldsymbol{\mu}_{\text{min}},\boldsymbol{\mu}_{\text{max}}]^{d}$. It has an internal state $\mathbf{s}_{t}$ and produces observations $\mathbf{o}_{t} \in \mathbb{R}^n$ at each time step $t \in \{1,\ldots,T\}$.
The environment interacts with a trained agent $\mathbf{A}$, which gets $\mathbf{o}_{t}$ from $\mathbf{E}_{\boldsymbol{\mu}}$ and produces actions $\mathbf{a}_{t}$. At each time step $t$ and after the agent performs $\mathbf{a}_{t}$, the internal state of the environment is updated:  $\mathbf{s}_{t+1} = \mathbf{E}_{\boldsymbol{\mu}}(\mathbf{s}_{t},\mathbf{a}_{t})$. The environment rewards the agent with $r_{t}=\mathit{R(\mathbf{s}_t, \mathbf{a}_t)}$, for some reward function $\mathit{R}$. We define the episode score $\mathit{Q}(\mathbf{A},\mathbf{E}_{\boldsymbol{\mu}})=\sum_{t=1}^{T}r_{t}$ of all intermediate rewards. The goal of $\mathbf{A}$ is to complete a task by maximizing $\mathit{Q}$. 
The adversary $\mathbf{G}$ attacks the agent $\mathbf{A}$ by modifying the environment $\mathbf{E}_{\boldsymbol{\mu}}$ through its parameters $\boldsymbol{\mu}$ without access to $\mathbf{A}$ and $\mathbf{E}_{\boldsymbol{\mu}}$.

\mysection{Distribution of Adversarial Attacks} \label{attack-distribution}
We define $\mathbf{P}_{\boldsymbol{\mu'}}$ to be the \textit{fooling distribution} of semantic parameters $\boldsymbol{\mu'}$ representing the environments $\mathbf{E}_{\boldsymbol{\mu}'}$, which fool the agent $\mathbf{A}$. 
\begin{equation}
\begin{aligned} 
 \boldsymbol{\mu'} \sim \mathbf{P}_{\boldsymbol{\mu'}}
 \Leftrightarrow \mathit{Q}(\mathbf{A},\mathbf{E}_{\boldsymbol{\mu'}}) \le \epsilon;~~ \boldsymbol{\mu'} ~\in ~[\boldsymbol{\mu}_{\text{min}},\boldsymbol{\mu}_{\text{max}}]^{d}
\label{eq:fool-distribution}
\end{aligned}
\end{equation}
Here, $\epsilon$ is a task-specific threshold to determine success and failure of the agent $\mathbf{A}$. The distribution $\mathbf{P}_{\boldsymbol{\mu'}}$ covers all samples that result in failure of $\mathbf{A}$. 
Its PDF is unstructured and depends on the complexity of the agent. 
We seek an adversary $\mathbf{G}$ that learns $\mathbf{P}_{\boldsymbol{\mu'}}$, so it can be used to comprehensively analyze the weaknesses of $\mathbf{A}$. Unlike the common practice of finding adversarial examples (\eg individual images), we address the attacks distribution-wise in a compact semantic parameter space. 
We denote our analysis technique as Semantic Adversarial Diagnostic Attack (SADA). \emph{Semantic} because of the nature of the environment parameters and \emph{diagnostic} because a fooling distribution is sought. We show later how this distribution can be used to reveal agents' failure modes. 
We propose to optimize the following objective for the adversary $\mathbf{G}$ to achieve this challenging goal:
\begin{equation}
\begin{aligned} 
 &\argmin_{\mathbf{G}} ~~ \mathbb{E}_{\boldsymbol{\mu}\sim \mathbf{G}} [\mathit{Q}(\mathbf{A},\mathbf{E}_{\boldsymbol{\mu}})]  \\
  & \text{s.t.}~~\{\boldsymbol{\mu}: \boldsymbol{\mu} \sim \mathbf{G}\} =  \{\boldsymbol{\mu'}: \boldsymbol{\mu'} \sim \mathbf{P}_{\boldsymbol{\mu'}} \}
\label{eq:objective}
\end{aligned}
\end{equation}

\begin{algorithm}[!t] 
\caption{Generic Adversarial Attacks on Agents}\label{alg: attacks}
\small
\SetAlgoLined
  \textbf{Returns: }Attack fooling Rate (AFR)\\
  \textbf{Requires: } Agent $\mathbf{A}$, Adversary $\mathbf{G}$, Environment $\mathbf{E}_{\boldsymbol{\mu}}$, number of episodes $T$, training iterations $L$, test size $M$, fooling threshold $\epsilon$ \\
  \textbf{Training $\mathbf{G}$: }
  \For{$i \leftarrow 1$ \KwTo $L$}{
   Sample $\boldsymbol{\mu}_{i} \sim \mathbf{G}$ and 
   initialize $\mathit{E_{\boldsymbol{\mu}_i}}$ with initial state $\mathbf{s}_{1}$ \\
   \For{$t \leftarrow 1$ \KwTo $T$}{
   $\mathit{E_{\boldsymbol{\mu}_i}}$ produces observation $\mathbf{o_t}$ from $\mathbf{s_t}$ \\
    $\mathbf{A}$ performs  $\mathbf{a}_{t}(\mathbf{o_t})$ and  receives   $r_{t} \leftarrow \mathit{R}(\mathbf{s}_{t}, \mathbf{a}_{t})$ \\
    State updates: $\mathbf{s}_{t+1} \leftarrow \mathit{E_{\boldsymbol{\mu}_i}}(\mathbf{s}_{t},\mathbf{a}_{t})$
   }
   $\mathbf{G}$ receives the episode score $\mathit{Q}_{i}(\mathbf{A},\mathbf{E}_{\boldsymbol{\mu}}) \leftarrow \sum_{t=1}^{T}r_{t}$ \\
   Update $\mathbf{G}$ to solve for \eqLabel{\ref{eq:objective}} \\
 }
 \textbf{Testing $\mathbf{G}$: }
    Initialize fooling counter $f \leftarrow$ 0 \\
   \For{$j \leftarrow 1$ \KwTo $M$}{
    sample $\boldsymbol{\mu}_{j} \sim \mathbf{G}$ and initialize $\mathit{E_{\boldsymbol{\mu}_j}}$ with initial state $\mathbf{s}_{1}$ \\
   \For{$t \leftarrow 1$ \KwTo $T$}{
    $\mathbf{a}_{t}(\mathbf{o_t})$ ; $r_{t} \leftarrow \mathit{R}(\mathbf{s}_{t}, \mathbf{a}_{t})$ ~;~ $\mathbf{s}_{t+1} \leftarrow \mathit{E_{\boldsymbol{\mu}_j}}(\mathbf{s}_{t},\mathbf{a}_{t})$ \\
    }
    $\mathit{Q}_{j}(\mathbf{A},\mathbf{E}_{\boldsymbol{\mu}_j}) \leftarrow \sum_{t=1}^{T}r_{t}$\\
    \If{$\mathit{Q}_{j}(\mathbf{A},\mathbf{E}_{\boldsymbol{\mu}_j}) ~ \leq \epsilon~$}
    {$f ~\leftarrow~ f + 1$   
    }
    }
    \textbf{Returns: }AFR = $f/M$
\end{algorithm}
\begin{figure*}[t]
\centering
   \includegraphics[width=0.95\linewidth]{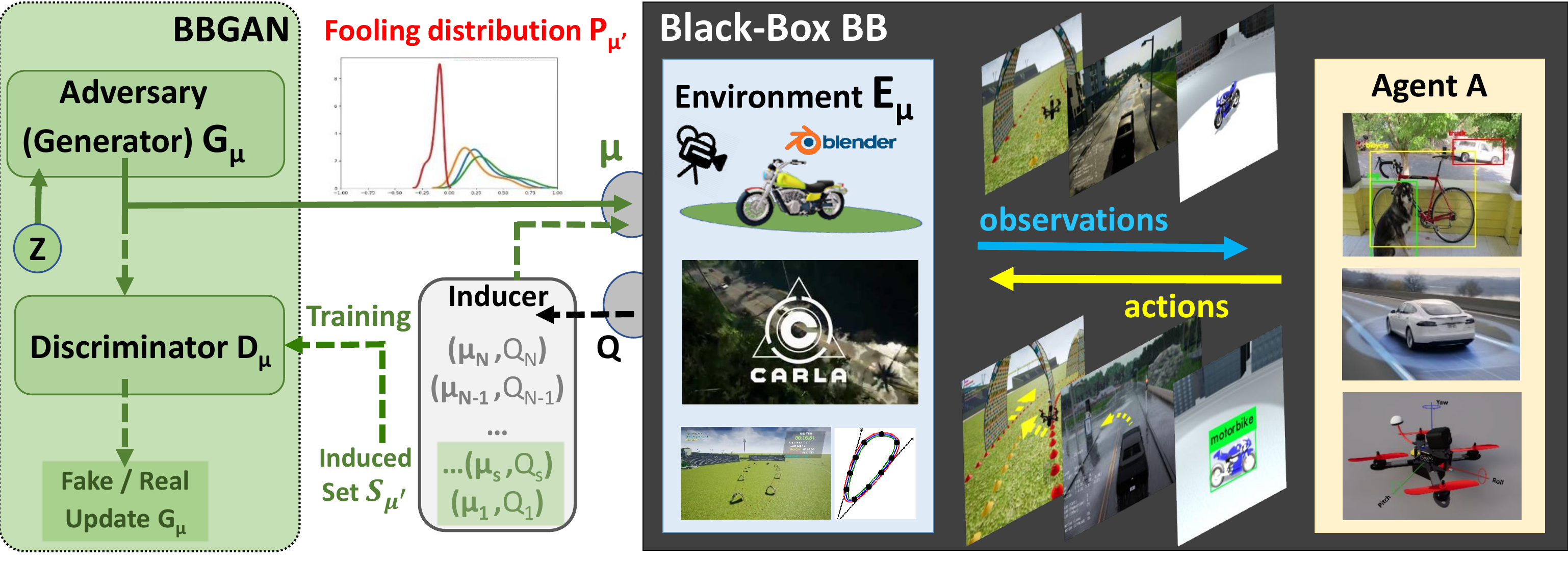}
   \caption{\small \textbf{BBGAN: Learning Fooling Distribution of Semantic Environment Parameters}. We learn an adversary $\mathbf{G}$, which samples semantic parameters $\boldsymbol{\mu}$ that parametrize the environment $\mathbf{E}_{\boldsymbol{\mu}}$, such that an agent $\mathbf{A}$ fails in a given task in $\mathbf{E}_{\boldsymbol{\mu}}$. The inducer produces the induced set $S_{\boldsymbol{\mu'}}$ from a uniformly sampled set $\Omega$ by filtering the lowest scoring $\boldsymbol{\mu}$ (according to $Q$ value), and passing $S_{\boldsymbol{\mu'}}$ for BBGAN training. Note that $\mathit{Q}_{1}\le\mathit{Q}_{s}...,~\le\mathit{Q}_{N}$, where $s = \left| S_{\boldsymbol{\mu'}}\right|,~ N = \left| \Omega\right| $. The inducer and the discriminator are only used during training (dashed lines), after which the adversary learns the fooling distribution $\mathbf{P}_{\boldsymbol{\mu'}}$. Three safety-critical applications are used to demonstrate this in three virtual environments: object detection (in Blender \cite{blender}), self-driving cars (in CARLA \cite{carla}), and autonomous racing UAVs (in Sim4CV \cite{sim4cv}).}
   \label{fig:pipeline}
\end{figure*}
Algorithm \ref{alg: attacks} describes a general setup for $\mathbf{G}$ to learn to generate fooling parameters. It also includes a mechanism for evaluating $\mathbf{G}$ in the black-box environment $\mathbf{E}_{\boldsymbol{\mu}}$ for $L$ iterations after training it to attack the agent $\mathbf{A}$. An attack is considered a fooling attack, if parameter $\boldsymbol{\mu}$ sampled from $\mathbf{G}$ achieves an episode score $\mathit{Q}(\mathbf{A},\mathbf{E}_{\boldsymbol{\mu}})\le \epsilon$. Consequently, the Attack Fooling Rate (AFR) is defined as the rate at which samples from $\mathbf{G}$ are fooling attacks. In addition to AFR, the algorithm returns the set $S_{\boldsymbol{\mu'}}$ of adversarial examples that can be used to diagnose the agent. The equality constraint in \eqLabel{\ref{eq:objective}} is very strict to include \textit{all} fooling parameters $\boldsymbol{\mu'}$ of the fooling distribution. It acts as a perceptuality metric in our generalized attack to prevent unrealistic attacks. Next, we relax this equality constraint to leverage recent advances in GANs for learning an estimate of the distribution $\mathbf{P}_{\boldsymbol{\mu'}}$.

\subsection{Black-Box Generative Adversarial Network} \label{BBGAN}
Generative Adversarial Networks (GANs) are a promising family of unsupervised techniques that can model complex domains, \eg natural images \cite{GAN,WGAN,Bigan}. GANs consist of a discriminator $\mathbf{D}_{\mathbf{x}}$ and a generator $\mathbf{G}_{\mathbf{x}}$ that are adversarially trained to optimize the loss $\mathbf{\mathit{L}}_{GAN}(\mathbf{G}_{\mathbf{x}},\mathbf{D}_{\mathbf{x}},\mathbf{P}_{X})$, where $\mathbf{P}_{X}$ is the distribution of images in domain X and $ \mathbf{z} \in \mathbb{R}^{c}$ is a latent random Gaussian vector.
\begin{align}
&\min_{\mathbf{G}_{\mathbf{x}}} \max_{\mathbf{D}_{\mathbf{x}}}~~\mathbf{\mathit{L}}_{\text{GAN}}(\mathbf{G}_{\mathbf{x}},\mathbf{D}_{\mathbf{x}},\mathbf{P}_{X})=  \label{eq:GAN} \\
& \mathbb{E}_{\mathbf{x}\sim p_{x}(\mathbf{x})} [\log \mathbf{D}_{\mathbf{x}}(\mathbf{x})] +  \mathbb{E}_{\mathbf{z}\sim p_{\mathbf{z}}(\mathbf{z})} [\log (1- \mathbf{D}_{\mathbf{x}}(\mathbf{G}_{\mathbf{x}}(\mathbf{z})))] \notag
\end{align}

$\mathbf{D}_{\mathbf{x}}$ tries to determine if a given sample (\eg image $\mathbf{x}$) is real (exists in the training dataset) or fake (generated by $\mathbf{G}_{\mathbf{x}}$). On the other hand, $\mathbf{G_{\mathbf{x}}}$ tries to generate samples that fool $\mathbf{D_{\mathbf{x}}}$ (\eg misclassification). Both networks are proven to converge when $\mathbf{G_{\mathbf{x}}}$ can reliably produce the underlying distribution of the real samples \cite{GAN}.

We propose to learn the fooling distribution $\mathbf{P}_{\boldsymbol{\mu'}}$ using a GAN setup, which we denote as black-box GAN (BBGAN).
We follow a similar GAN objective but replace the image domain $\mathbf{x}$ by the semantic environment parameter $\boldsymbol{\mu}$. However, since we do not have direct access to 
$\mathbf{P}_{\boldsymbol{\mu'}}$, 
we propose a module called the \textit{inducer}, which is tasked to produce the \textit{induced set} $S_{\boldsymbol{\mu'}}$ that belongs to $\mathbf{P}_{\boldsymbol{\mu'}}$. In essence, the \textit{inducer} tries to choose a parameter set which represents the fooling distribution to be learnt by the BBGAN as well as possible. In practice, the inducer selects the best fooling parameters (based on the $Q$ scores of the agent under these parameters) from a set of randomly sampled parameters in order to construct this \textit{induced set}. Thus, this setup relaxes \eqLabel{\ref{eq:objective}} to:
\begin{equation}
\begin{aligned} 
 &\argmin_{\mathbf{G}} ~~ \mathbb{E}_{\boldsymbol{\mu}\sim \mathbf{G}} [\mathit{Q}(\mathbf{A},\mathbf{E}_{\boldsymbol{\mu}})]  \\
  & \text{s.t.}~~\{\boldsymbol{\mu}: \boldsymbol{\mu} \sim \mathbf{G}\} \subset \{\boldsymbol{\mu'}: \boldsymbol{\mu'} \sim \mathbf{P}_{\boldsymbol{\mu'}} \}
\label{eq:relax}
\end{aligned}
\end{equation}
\noindent So, the final BBGAN loss becomes:
\begin{equation} 
\begin{aligned}
&
\min_{\mathbf{G}_{\boldsymbol{\mu}}} \max_{\mathbf{D}_{\boldsymbol{\mu}}}~~\mathbf{\mathit{L}}_{\text{BBGAN}}(\mathbf{G}_{\boldsymbol{\mu}},\mathbf{D}_{\boldsymbol{\mu}},S_{\boldsymbol{\mu'}})=  \\
& \mathbb{E}_{\boldsymbol{\mu}\sim S_{\boldsymbol{\mu'}}} [\log \mathbf{D}_{\boldsymbol{\mu}}(\boldsymbol{\mu})] +  \mathbb{E}_{\mathbf{z}\sim p_{\mathbf{z}}(\mathbf{z})} [\log (1- \mathbf{D}(\mathbf{G}(z)))]
\end{aligned}
\label{eq:BBGAN}
\end{equation}

\begin{table*}[!htb]
\footnotesize
\setlength{\tabcolsep}{8pt} 
\renewcommand{\arraystretch}{1.1} 
\centering
\begin{tabular}{c|c|c|c} 
\toprule
\textbf{Attack Variables} & \textbf{\specialcell{ Pixel Adversarial Attack\\ on Image Classifiers}} & \textbf{\specialcell{Semantic Adversarial Attack\\ on Object Detectors}} & \textbf{\specialcell{ Semantic Adversarial Attack\\ on Autonomous Agents}} \\
\midrule
Agent $\mathbf{A}$	& \specialcell{K-class classifier \\$\mathbf{C}:[0,1]^{n} \rightarrow [l_{1}, l_{2}, ... ,l_{K}]$ \\ $l_{j}$ : the softmax value for class $j$ } &  \specialcell{K-class object detector \\ $\mathbf{F}:[0,1]^{n} \rightarrow  (\mathbb{R}^{N\times K},\mathbb{R}^{N\times 4})$ \\ $N$ : number of detected objects }	& \specialcell{self-driving policy agent $\mathbf{A}$\\ \eg network to regress controls } \\ \hline
Parameters $\boldsymbol{\mu}$ & \specialcell{ the pixels noise \\ added on attacked image $\mathbf{x}_{i}$}  & \specialcell{parameters describing the scene\\\eg camera pose, object , light}  & \specialcell{parameters involved in the simulation\\ \eg road shape , weather , camera}\\ \hline
Environment $\mathbf{E}_{\boldsymbol{\mu}}$ & \specialcell{ dataset $\Phi$ containing all images\\ and their true class label \\ $\Phi = \{(\mathbf{x}_{i}, y_{i} )\}_{i=1}^{|\Phi|}$} &  \specialcell{ dataset $\Phi$ containing all\\ images and their true\\  class label} & \specialcell{ simulation environment \\ partially described by $\boldsymbol{\mu}$\\ that $\mathbf{A}$ navigates in for a target}\\ \hline
Observation $\mathbf{o}_{t}$ &\specialcell{attacked image after added noise\\= $\mathbf{x}_{i} + \boldsymbol{\mu}$,  where  $\mathbf{x} ~,~ \boldsymbol{\mu} \in \mathbb{R}^{n}$ } & \specialcell{the rendered image \\using the scene parameters $\boldsymbol{\mu}$ }  & \specialcell{sequence of rendered images $\mathbf{A}$\\observes during the simulation episode}\\ \hline
Agent actions $\mathbf{a}_{t}(\mathbf{o}_{t})$ & \specialcell{predicted softmax vector \\ of attacked image}  &  \specialcell{predicted confidence of \\ the true class label} & \specialcell{steering command to move\\ the car/UAV in the next step} \\ \hline
Score $\mathit{Q}(\mathbf{A},\mathbf{E}_{\boldsymbol{\mu}})$ & \specialcell{the difference between \\true and predicted softmax} &    \specialcell{predicted confidence of \\ the true class label} & \specialcell{the average sum of rewards \\ over five different episodes}  \\ 
\bottomrule
\end{tabular}
\caption{\small\textbf{Cases of Generic Adversarial Attacks}: variable substitutions that lead to known attacks.}
\label{tbl:var-summary}
\end{table*}

Here, $\mathbf{G}_{\boldsymbol{\mu}}$ is the generator acting as the adversary, and $\mathbf{z}\in \mathbb{R}^{m}$ is a random variable sampled from a normal distribution. A simple inducer can be just a filter that takes a uniformly sampled set $\Omega = \{\boldsymbol{\mu}_{i} \sim \text{Uni}([\boldsymbol{\mu}_{\text{min}},\boldsymbol{\mu}_{\text{max}}])\}_{i=1}^{i=N}$ and suggests the lowest $\mathit{Q}$-scoring $\boldsymbol{\mu}_{i}$ that satisfies the condition $\mathit{Q}(\boldsymbol{\mu}_{i}) \le \epsilon$. The selected samples constitute the induced set $S_{\boldsymbol{\mu'}}$.
The BBGAN treats the induced set as a training set, so the samples in $S_{\boldsymbol{\mu'}}$ act as virtual samples from the fooling distribution $\mathbf{P}_{\boldsymbol{\mu'}}$ that we want to learn. As the induced set size $S_{\boldsymbol{\mu'}}$ increases, the BBGAN learns more of 
$\mathbf{P}_{\boldsymbol{\mu'}}$. As $|S_{\boldsymbol{\mu'}}| \rightarrow \infty , $ any sample from $S_{\boldsymbol{\mu'}}$ is a sample of $\mathbf{P}_{\boldsymbol{\mu'}}$ and the BBGAN in \eqLabel{\ref{eq:BBGAN}} satisfies the strict condition in \eqLabel{\ref{eq:objective}}. Consequently, sampling from $\mathbf{G}_{\boldsymbol{\mu}}$ would consistently fool agent $\mathbf{A}$. We show an empirical proof for this in the \supp and show how we consistently fool three different agents by samples from $\mathbf{G}_{\boldsymbol{\mu}}$ in the experiments. The number of samples needed for $S_{\boldsymbol{\mu'}}$ to be representative of $\mathbf{P}_{\boldsymbol{\mu'}}$ depends on the 
dimensionality $d$ of $\boldsymbol{\mu}$. Because of the black-box and stochastic nature of $\mathbf{E}_{\boldsymbol{\mu}}$ and $\mathbf{A}$ (similar to other RL environments), we follow the random sampling scheme common in RL \cite{random-rl} instead of deterministic gradient estimation. 
In the experiments, we compare our method against baselines that use different approaches to solve \eqLabel{\ref{eq:objective}}. 

\subsection{Special Cases of Adversarial Attacks} \label{sec:special-cases}
One can show that the generic adversarial attack framework detailed above includes well-known types of attacks as special cases, summarized in Table \ref{tbl:var-summary}. 
In fact, the general setup allows for static agents (\eg classifiers and detectors) as well as dynamic agents (\eg an autonomous agent acting in a dynamic environment). It also covers pixel-wise image perturbations, as well as, semantic attacks that try to fool the agent in a more realistic scenario. The generic attack also allows for a more flexible way to define the attack success based on an application-specific threshold $\epsilon$ and the agent score $\mathit{Q}$.
In the \supp, we provide more details on the inclusiveness of our generic setup and how it covers original pixel-level adversarial attack objectives.

\section{Applications}\label{sec:applications}
\subsection{Object Detection} \label{sec:application-detection}
\begin{figure}[!t]
\centering
\includegraphics[width=1\columnwidth]{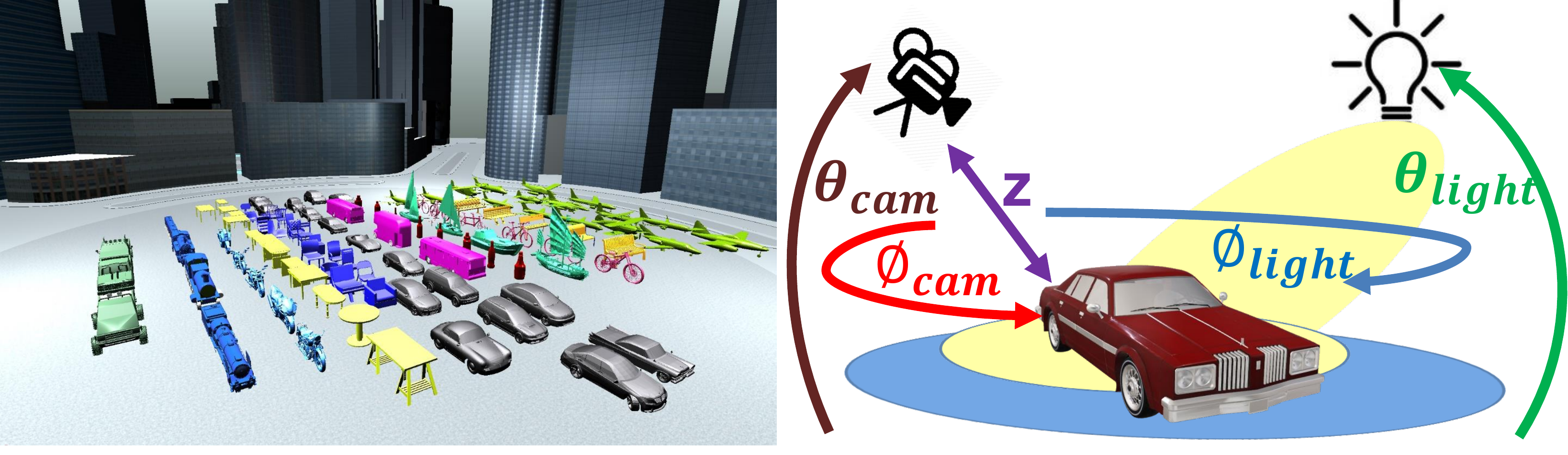}
\caption{\small \textbf{Object Detection Attack Setup}: (\emph{Left}): the 100 shapes from Pascal3D  and ShapeNet of 12 object classes, used to uncover the failure cases of the YOLOV3 detector.
(\emph{Right}): the semantic parameters $\boldsymbol{\mu}$ defining the environment. ($z$): camera distance to the object, ($\phi_{\text{cam}},\theta_{\text{cam}},\phi_{\text{light}},\phi_{\text{light}}$): camera azimuth, pitch and light source azimuth, and pitch angles respectively. }
\label{fig:parameters}
\end{figure}

Object detection is one of the core perception tasks commonly used in autonomous navigation. 
Based on its suitability for autonomous applications, we choose the very fast, state-of-the-art YOLOv3 object detector \cite{yolo3} as the agent in our SADA framework. 
We use the open-source software Blender to construct a scene based on freely available 3D scenes and CAD models. We pick an urban scene with an open area to allow for different rendering setups. The scene includes one object of interest as well as a camera and main light source directed toward the center of the object. The light is a fixed strength spotlight located at a fixed distance from the object. The material of each object is semi-metallic, which is common for the classes under consideration. 
The 3D collection consists of 100 shapes of 12 object classes (aeroplane, bench, bicycle, boat, bottle, bus, car, chair, dining table, motorbike, train, truck) from Pascal-3D \cite{pascal3D} and ShapeNet \cite{shapenet}.
At each iteration, one shape from the intended class is randomly picked and placed in the middle of the scene. The rendered image is then passed to YOLOV3 for detection. 
For the environment parameters, we use eight parameters that have shown to affect detection performance and frequently occur in real setups (refer to \figLabel \ref{fig:parameters}). 

\subsection{Self-Driving} \label{selfdrive}
There is a lot of recent work in autonomous driving especially in the fields of robotics and computer vision \cite{Franke2017,Codevilla2018}. In general, complete driving systems are very complex and difficult to analyze or simulate. By learning the underlying distribution of failure cases, our work provides a safe way to analyze the robustness of such a complete system. While our analysis is done in simulation only, we would like to highlight that sim-to-real transfer is a very active research field nowadays
\cite{Sadeghi2017,Tobin2017}.
We use an autonomous driving agent (based on CIL \cite{Codevilla2018}), which was trained on the environment $\mathbf{E}_{\mu}$ with default parameters. The driving-policy was trained end-to-end to predict car controls given an input image and is conditioned on high-level commands (\eg \emph{turn right at the next intersection}).
The environment used is CARLA driving simulator \cite{carla}, the most realistic open-source urban driving simulator currently available. We consider the three common tasks of driving in a straight line, completing one turn, and navigating between two random points. The score is measured as the average success of five pairs of start and end positions.
Since experiments are time-consuming, we restrict ourselves to three parameters, two of which pertain to the mounted camera viewpoint and the third controls the appearance of the environment by changing the weather setting (\eg 'clear noon', 'clear sunset', 'cloudy after rain', etc.). As such, we construct an environment by randomly perturbing the position and rotation of the default camera along the z-axis and around the pitch axis respectively, and by picking one of the weather conditions. Intuitively, this helps measure the robustness of the driving policy to the camera position (\eg deploying the same policy in a different vehicle) and to environmental conditions. 

\begin{figure} [!t] 
\tabcolsep=0.03cm
\begin{tabular}{ccc}  
\includegraphics[width = \columnwidth]{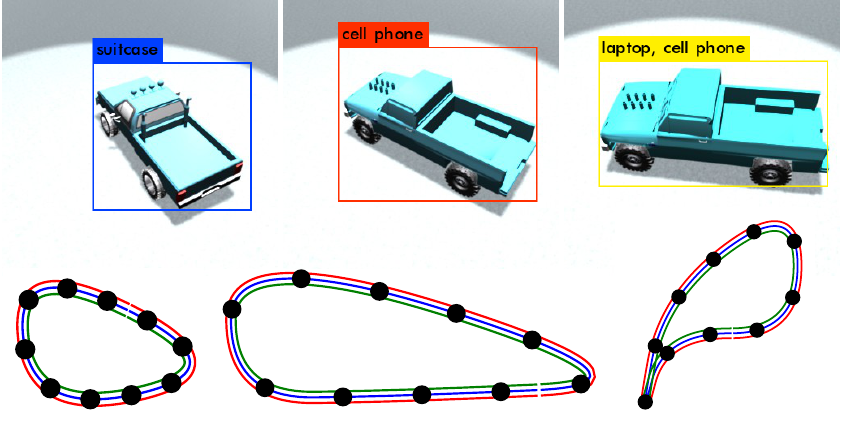} &
\end{tabular}
\caption{\small \textbf{Qualitative Examples}: (\emph{Top}): BBGAN generated samples that fool YOLOV3 detector on the Truck class. (\emph{Bottom}): BBGAN generated tracks that fool the UAV navigation agent. 
}
\label{fig:qualitative}
\end{figure}

\begin{table*}[t]
\footnotesize
\centering
\setlength{\tabcolsep}{4pt} 
\renewcommand{\arraystretch}{0.95} 
\begin{tabular}{lcccc|ccc|ccc}
\toprule
& \multicolumn{4}{c}{Object Detection} & \multicolumn{3}{c}{Autonomous Driving} & \multicolumn{3}{c}{UAV Track Generation} \\
& Bicycle & Motorbike & Truck & 12-class avg & Straight  & One Curve  & Navigation  & 3 anchors & 4 anchors & 5 anchors  \\
\midrule
Full Set                                  & 14.6\% & 32.5\%    & 56.8\%   & 37.1 \%   & 10.6\%     & 19.5\%     & 46.3\%      & 17.0\%       & 23.5\%      & 15.8\%      \\
Random                                    & 13.3\% & 38.8\%    & 73.8\%   & 45.7\%    & 8.0\%      & 18.0\%     & 48.0\%      & 22.0\%       & 30.0\%      & 16.0\%      \\
Multi-Class SVM                           & 20.0\% & 45.6\%    & 70.8\%   & 45.8\%    & 96.0\%     & \textbf{100}\%      & \textbf{100}\%       & 24.0\%       & 30.0\%      & 14.0\%      \\
GP Regression                             & 17.6\% & 43.6\%    & 83.6\%    & 45.26\%   & \textbf{100}\%      & \textbf{100}\%      & \textbf{100}\%       & 74.0\%       & 94.0\%      & 44.0\%      \\
Gaussian                                  & 19.6\% & 40.4\% & 72.4\%   & 47.0\%    & 54.0\%     & 30.0\%     & 64.0\%      & 49.3\%       & 56.0\%      & 28.7\%      \\
GMM10\%                                   & 26.0\% & 48.4\% & 75.2\%   & 49.0\%    & 90.0\%     & 72.0\%     & 98.0\%      & 57.0\%       & 63.0\%      & 33.0\%      \\
GMM50\%                                   & 16.4\% & 46.8\% & 72.0\%  & 47.8\%    & 92.0\%     & 68.0\%     & \textbf{100}\%       & 54.0\%       & 60.0\%      & 40.0\%      \\
Bayesian                   & 48.0\% & 52.0\% & 75.6\%   & 56.1\%    & -          & -          & -           & -            & -           & -           \\
\midrule
BBGAN (vanilla)                              & 44.0\% & 45.2\% & 90.8\%   & 74.5\%    & \textbf{100}\%      & 98.0\%     & 98.0\%           & 42.0\%       & 94.0\%       & 86.0\%      \\
\textbf{BBGAN (boost)}                             & \textbf{65.8}\% & \textbf{82.0}\% & \textbf{100}\% & \textbf{80.5}\%    & \textbf{100}\%          & \textbf{100}\%          & \textbf{100}\%           & \textbf{86.0}\%            & \textbf{98.0}\%           & \textbf{92.0}\%           \\
\bottomrule
\end{tabular}
\caption{\small \textbf{Attack Fooling Rate (AFR) Comparison:} AFR of adversarial samples generated on three safety-critical applications: YOLOV3 object detection, self-driving, and UAV racing. 
For detection, we report the average AFR performance across all 12 classes and highlight 3 specific ones.
For autonomous driving, we compute  the AFR for the three common tasks in CARLA. For UAV racing, we compute AFR for race tracks of varying complexity (3, 4, or 5 anchor points describe the track).  
We see that our BBGAN outperforms all the baselines, and with larger margins for higher dimensional tasks (\eg detection). Due to the expensive computations and sequential nature of the Bayesian baseline, we omit it for the two autonomous navigation applications. Best results are in \textbf{bold}.
}
\label{tbl:all}
\end{table*}
\subsection{UAV Racing} \label{racingtrack}
In recent years, UAV (unmanned aerial vehicle) racing has emerged as a new sport where pilots compete in navigating small UAVs through race courses at high speeds. Since this is a very interesting research problem, it has also been picked up by the robotics and vision communities \cite{pmlr-v87-kaufmann18a}. 
We use a fixed agent to autonomously fly through each course and measure its success as percentage of gates passed \cite{TeachingUAVstoRace}. If the next gate was not reached within 10 seconds, we reset the agent at the last gate. We also record the time needed to complete the course. The agent uses a perception network that produces waypoints from image input and a PID controller to produce low-level controls. 
We use the general-purpose simulator for computer vision applications, Sim4CV \cite{sim4cv}.
Here, we change the geometry of the race course environment rather than its appearance. We define three different race track templates with 3-5 2D anchor points, respectively. These points describe a second order B-spline and are perturbed to generate various race tracks populated by uniformly spaced gates. Please refer to the \supp for more details.

\section{Experiments} \label{sec:experiments}
\subsection{BBGAN}
\mysection{Training}\label{training}
To learn the fooling distribution $\mathbf{P}_{\boldsymbol{\mu'}}$, we train the BBGAN using a vanilla GAN model \cite{GAN}. Both, the generator $\mathbf{G}$ and the discriminator $\mathbf{D}$ consist of a MLP with 2 layers. We train the GAN following convention, but since we do not have access to the true distribution that we want to learn (\ie real samples), we \emph{induce} the set by randomly sampling $N$ parameter vector samples $\boldsymbol{\mu}$, and then picking the $s$ worst  among them (according to the score $Q$). For object detection, we use $N=20000$ image renderings for each class (a total of 240K images).
Due to the computational cost, our dataset for the autonomous navigation tasks comprises only $N=1000$ samples. For instance, to compute one data point in autonomous driving, we need to run a complete episode that requires 15 minutes. The induced set size is always fixed to be $s=100$. 

\mysection{Boosting}
We use a boosting strategy to improve the performance of our BBGAN. Our boosting strategy simply utilizes the samples generated by the previous stage adversary $\mathbf{G}_{k-1}$ in inducing the training set for the current stage adversary $\mathbf{G}_{k}$. This is done by adding the generated samples to $\Omega$ before training $\mathbf{G}_{k}$. The intuition here is that the main computational burden in training the BBGAN is not the GAN training itself, but computing the agent episodes, each of which can take multiple hours for the case of self-driving. For more details, including the algorithm, a mathematical justification and more experimental results please refer to the \supp. 
\subsection{Testing, Evaluation, and Baselines}
To highlight the merits of BBGAN, we seek to compare it against baseline methods, which also aim to estimate the fooling distribution $\mathbf{P}_{\boldsymbol{\mu'}}$. In this comparative study, each method 
produces $M$ fooling/adversarial samples (250 for object detection and 100 for self-driving and UAV racing) based on its estimate of $\mathbf{P}_{\boldsymbol{\mu'}}$. Then, the \textit{attack fooling rate} (AFR) for each method is computed as the percentage of the  $M$ adversarial samples that fooled the agent. To determine whether the agent is fooled, we use a fooling rate threshold  $\epsilon = 0.3$ \cite{shapeshifter}, $\epsilon = 0.6$, and $\epsilon = 0.7$ for object detection, self-driving, and UAV racing, respectively. In the following, we briefly explain the baselines. 
\mysection{Random} We uniformly sample random parameters $\boldsymbol{\mu}$ within an admissible range that is application dependent.

\mysection{Gaussian Mixture Model (GMM)} We fit a full covariance GMM of varying Gaussian components to estimate the distribution of the samples in the induced set $S_{\boldsymbol{\mu'}}$. The variants are denoted as Gaussian (one component), GMM10\% and GMM50\% (number of components as percentage of the samples in the induced set).

\mysection{Bayesian} We use the Expected Improvement (EI) Bayesian Optimization algorithm \cite{expected-improvment} to minimize the score $\mathit{Q}$ for the agent. The optimizer runs for 10$^4$ steps and it tends to gradually sample more around the global minimum of the function. We use the last $N=1000$ samples to generate the induced set $S_{\boldsymbol{\mu'}}$ and then learn a GMM with different Gaussian components. Finally, we sample $M$ parameter vectors from the GMMs and report results for the best model.

\mysection{Multi-Class SVM}
We bin the score $Q$ into 5 equally sized bins and train a multi-class SVM classifier on the complete set $\Omega$ to predict the correct bin.
We then randomly sample parameter vectors $\boldsymbol{\mu}$, classify them, and sort them by the predicted score. We pick $M$ samples with the lowest $Q$ score.

\mysection{Gaussian Process Regression}
Similar to the SVM case, we train a Gaussian Process Regressor \cite{gp-robot} with an exponential kernel to regress the scores $Q$ from the corresponding $\boldsymbol{\mu}$ parameters that generated the environment on the dataset $\Omega$. 

\begin{figure}[!t]
\includegraphics[width=\columnwidth]{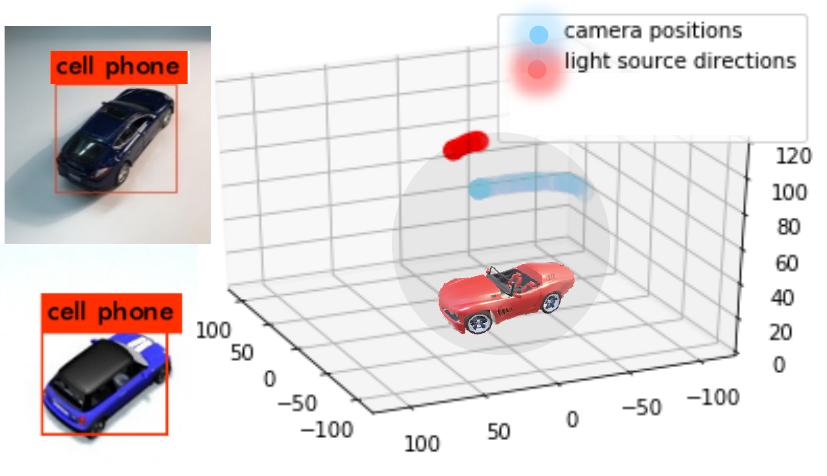}
\caption{\small \textbf{Visualization of the Fooling Distribution}. 
(\textit{right}): We plot the camera positions and light source directions of 250 sampled parameters in a 3D sphere around the object. (\textit{left}): We show how real photos of a toy car, captured from the same angles as rendered images, confuse the YOLOV3 detector in the same way.}
\label{fig:analysis-car}
\end{figure}
\subsection{Results} \label{sec:Results}
Table \ref{tbl:all} summarizes the AFR results for the aforementioned baselines and our BBGAN approach across all three applications. For object detection, we show 3 out of 12 classes and report the average across all classes. For autonomous driving, we report the results on all three driving tasks. For UAV racing, we report the results for three different track types, parameterized by an increasing number of 2D anchor points (3, 4 and 5) representing $\boldsymbol{\mu}$. 
Our results show that we consistently outperform the baselines, even the ones that were trained on the complete set $\Omega$ rather than the smaller induced set $S_{\boldsymbol{\mu'}}$, such as the multi-class SVM and the GP regressor. While some baselines perform well on the autonomous driving application where $\boldsymbol{\mu}$ consists  of only 3 parameters, our approach outperforms them by a large margin on the tasks with higher dimensional $\boldsymbol{\mu}$ (\eg object detection and UAV racing with 5-anchor tracks). Our boosting strategy is very effective and improves results even further with diminishing returns for setups where the vanilla BBGAN already achieves a very high or even the maximum success rate.

To detect and prevent mode collapse, a GAN phenomenon where the generator collapses to generate a single point, we do the following. (1) We visualize the Nearest Neighbor (NN) of the generated parameters in the training set as in \figLabel\ref{fig:nn}. (2) We visualize the distributions of the generated samples and ensure their variety as in \figLabel\ref{fig:analysis-car}.
(3) We measure the average standard deviation per parameter dimension to make sure it is not zero.
(4) We visualize the images/tracks created by these parameters as in \figLabel\ref{fig:qualitative}. 

\label{sec:analysis}
\subsection{Analysis} 
\mysection{Diagnosis}
The usefulness of SADA lies in that it is not only an attacking scheme using BBGAN, but also serves as a diagnosis tool to assess the systematic failures of agents. We perform diagnostic studies (refer to \figLabel\ref{fig:analysis-car}) to identify cases of systematic failure for the YOLOv3 detector.

\begin{figure}[t]
\tabcolsep=0.03cm
\includegraphics[width = \columnwidth]{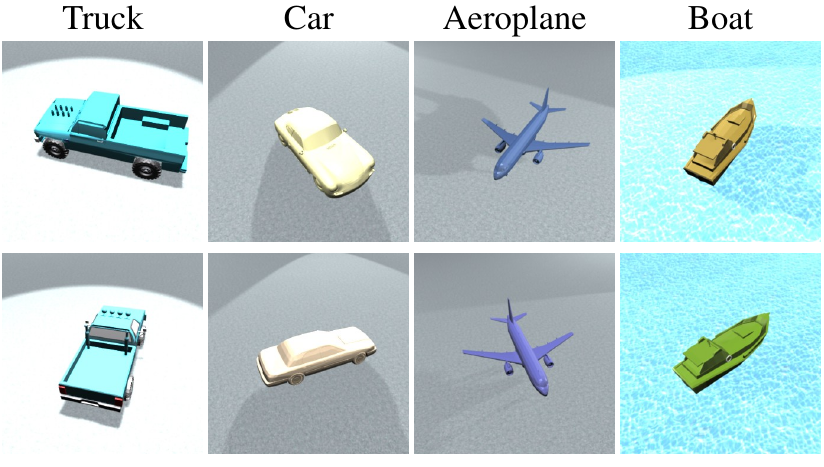}

\caption{\small \textbf{Nearest Neighbor in Training}: (\emph{top}): generated fooling samples by our BBGAN for 4 different classes. (\emph{bottom}): the corresponding NN from the training set. We can see that our BBGAN generates novel fooling parameters that are not present in training.}
\label{fig:nn}
\end{figure}

\mysection{Transferability} \label{sec:transferability}
To demonstrate the transferability of the fooling parameter distribution to the real-world, we photograph toy models using a standard mobile phone camera and use office desk lights analogous the light source of the virtual environment. We orient the camera and the lighting source according to the fooling distribution learned from the virtual world. In Figure \ref{fig:analysis-car}, we show a real-world photo of the toy car and the corresponding rendered virtual-world image, both of which are similarly fooled with the same erroneous class label. Please refer to the \supp for a similar analysis of other classes, the transfer of attacks between different CAD models and the effect of occlusion on detection. 

\mysection{Nearest Neighbor Visualization} 
In \figLabel\ref{fig:nn}, we visualize the NN in the parameter space for four different generated samples by our BBGAN. We see that the generated and the NN in training are different for the 4 samples with $L_{2}$ norm differences of (0.76,0.60,0.81,0.56) and (378,162,99,174) in parameter space and in image space respectively (all range from -1 to 1). This shows that our BBGAN can generate novel examples that fool the trained agent.

\section{Insights and Future Work}
\mysection{Object Detection with YOLOV3}
For most objects, top-rear or top-front views of the object tend to fool the YOLOV3 detector. The color of the object does not play a significant role in fooling the detector, but usually colors that are closer to the background color tend to be preferred by the BBGAN samples. 

\mysection{Self-Driving}
Weather is the least important parameter for fooling the driving policy indicating that the policy was trained to be insensitive to this factor. 
Interestingly, the learned policy is very sensitive to slight perturbations in the camera pose (height and pitch), indicating a systemic weakness that should be ratified with more robust training.

\mysection{UAV Autonomous Navigation}
We observe that the UAV fails if the track has very sharp turns. This makes intuitive sense and the results that were produced by our BBGAN consistently produce such tracks. While the tracks that are only parameterized by three control points can not achieve sharp turns, our BBGAN is still able to make the UAV agent fail by placing the racing gates very close to each other, thereby increasing the probability of hitting them. 

\mysection{Future Work}
Our work can be extended to other AI agents to test their semantic vulnerability to such attacks.  This can be used to establish more interpretable deep models and allow for safety-tests for AI models before deployment in real world safety-critical applications.   

\section{Acknowledgments}
This work was supported by the King Abdullah University of Science and Technology (KAUST) Office of Sponsored Research under Award No. RGC/3/3570-01-01.

{\small
\bibliographystyle{aaai}
\bibliography{egbib}
}
\clearpage
\appendix

\section{Empirical Justification for BBGAN }

We want to show that as the size of the induced set $|S_{\boldsymbol{\mu'}}| \rightarrow \infty$, learning an adversary according to the BBGAN objective in \eqLabel{\ref{sup:BBGAN}} converges to the fooling distribution of semantic parameters $\mathbf{P}_{\boldsymbol{\mu'}}$ defined in \eqLabel{\ref{sup:fool-distribution}}. 
\begin{equation} 
\begin{aligned}
&
\min_{\mathbf{G}_{\boldsymbol{\mu}}} \max_{\mathbf{D}_{\boldsymbol{\mu}}}~~\mathbf{\mathit{L}}_{\text{BBGAN}}(\mathbf{G}_{\boldsymbol{\mu}},\mathbf{D}_{\boldsymbol{\mu}},S_{\boldsymbol{\mu'}})=  \\
& \mathbb{E}_{\boldsymbol{\mu}\sim S_{\boldsymbol{\mu'}}} [\log \mathbf{D}_{\boldsymbol{\mu}}(\boldsymbol{\mu})] +  \mathbb{E}_{\mathbf{z}\sim p_{\mathbf{z}}(\mathbf{z})} [\log (1- \mathbf{D}(\mathbf{G}(z)))]
\end{aligned}
\label{sup:BBGAN}
\end{equation}

\begin{equation}
\begin{aligned} 
 \boldsymbol{\mu'} \sim \mathbf{P}_{\boldsymbol{\mu'}}
 \Leftrightarrow \mathit{Q}(\mathbf{A},\mathbf{E}_{\boldsymbol{\mu'}}) \le \epsilon;~~ \boldsymbol{\mu'} ~\in ~[\boldsymbol{\mu}_{\text{min}},\boldsymbol{\mu}_{\text{max}}]^{d}
\label{sup:fool-distribution}
\end{aligned}
\end{equation}

Here, the agent $\mathbf{A}$ interacts with the environment $\mathbf{E}_{\boldsymbol{\mu}}$ and receives a score $\mathit{Q}$ for a given parameter $\boldsymbol{\mu}$. The fooling threshold is $\epsilon$.

\subsection{Empirical Proof}
We use the definition of Exhaustive Search (Algorithm 3.1) from the Audet and Hare book on derivative-free and black box optimization \cite{bboptimization}. In this algorithm, we try to optimize an objective $\mathit{f}: \mathbb{R}^{d} \rightarrow \mathbb{R}$ defined on a closed continuous global set $\Omega$ by densely sampling a countable subset $\mathit{S} = \{\boldsymbol{\mu}_{1},\boldsymbol{\mu}_{2}, ..., \boldsymbol{\mu}_{N} \} \subset \Omega$. Theorem 3.1 \cite{bboptimization} states that as long as the exhaustive search continues infinitely from the set $\mathit{S}$, the global solutions of $\mathit{f}$ can be reached. Let's assume the global solutions $\boldsymbol{\mu}^{*}$ of $\mathit{f}$ exists and defined as follows.

\begin{equation}
\begin{aligned} 
 \boldsymbol{\mu}^{*}~ = ~ \argmin_{\boldsymbol{\mu}} ~~ \mathit{f}(\boldsymbol{\mu}) ~~~ s.t.~  \boldsymbol{\mu} \in \Omega 
\label{sup:theory}
\end{aligned}
\end{equation}
Let's denote the best solution reached up to the sample $\boldsymbol{\mu}_{N} $ to be $\boldsymbol{\mu}_{N}^{\text{best}}$ . If the set $S_{\boldsymbol{\mu}^{*}}$ is the set of all global solutions $\boldsymbol{\mu}^{*}$, then Theorem 3.1 \cite{bboptimization} states that
\begin{equation}
\begin{aligned} 
 \boldsymbol{\mu}_{N}^{\text{best}}~ \in ~ S_{\boldsymbol{\mu}^{*}} =\{ \boldsymbol{\mu}^{*} \} ,~~~ as~ N \rightarrow \infty 
\label{sup:theory2}
\end{aligned}
\end{equation}
Now let $\mathit{f}(\boldsymbol{\mu}) = \max(0,~ \mathit{Q}(\mathbf{E}_{\boldsymbol{\mu}},\mathbf{A}) - \epsilon )$ and let $\Omega = [\boldsymbol{\mu}_{\text{min}},\boldsymbol{\mu}_{\text{max}}]$, then the global solutions of the optimization:

\begin{equation}
\begin{aligned} 
 \boldsymbol{\mu}^{*}~ = ~ & \argmin_{\boldsymbol{\mu}} ~~ \max(0,~ \mathit{Q}(\mathbf{E}_{\boldsymbol{\mu}},\mathbf{A}) - \epsilon ) \\
 & s.t.~  \boldsymbol{\mu} \in [\boldsymbol{\mu}_{\text{min}},\boldsymbol{\mu}_{\text{max}}] 
\label{sup:infer}
\end{aligned}
\end{equation}
satisfy the two conditions in \eqLabel{\ref{sup:fool-distribution}} as follows.

\begin{equation}
\begin{aligned} 
 \mathit{Q}(\mathbf{A},\mathbf{E}_{\boldsymbol{\mu}^{*}}) \le \epsilon;~~ \boldsymbol{\mu}^{*} ~\in ~[\boldsymbol{\mu}_{\text{min}},\boldsymbol{\mu}_{\text{max}}]^{d}
\label{sup:satisfy}
\end{aligned}
\end{equation}

Hence, the set of all global solutions includes all the points in the fooling distribution:
\begin{equation}
\begin{aligned} 
  S_{\boldsymbol{\mu}^{*}} = \{ \boldsymbol{\mu}^{*} \} ~=  \{\boldsymbol{\mu'}: \boldsymbol{\mu'} \sim \mathbf{P}_{\boldsymbol{\mu'}} \}
\label{sup:objective}
\end{aligned}
\end{equation}

Therefore, as the sampling set size $|\mathit{S}| \rightarrow \infty $, all the points $\boldsymbol{\mu}$ that lead to $\mathit{Q}(\mathbf{E}_{\boldsymbol{\mu}},\mathbf{A}) \le \epsilon$, achieve the minimum objective in \eqLabel{\ref{sup:infer}} of zero and the set of best observed values $| \{\boldsymbol{\mu}_{N}^{\text{best}}\}| \rightarrow \infty  $. This set is what we refer to as the induced set $S_{\boldsymbol{\mu'}}$. From \eqLabel{\ref{sup:theory2}} and \eqLabel{\ref{sup:objective}}, we can infer that the induced set will include all fooling parameters as follows.

\begin{equation}
\begin{aligned} 
\text{as}~  N \rightarrow \infty,~ ~ S_{\boldsymbol{\mu'}} ~ \rightarrow  ~\{\boldsymbol{\mu'}: \boldsymbol{\mu'} \sim \mathbf{P}_{\boldsymbol{\mu'}} \}
\label{sup:objective2}
\end{aligned}
\end{equation}
Finally if the set $S_{\boldsymbol{\mu'}}$ has converged to the distribution $\mathbf{P}_{\boldsymbol{\mu'}}$ and we use $S_{\boldsymbol{\mu'}}$ to train the BBGAN in \eqLabel{\ref{sup:BBGAN}}, then according to proposition 2 from the original GAN paper by Goodfellow \etal \cite{GAN}, the adversary $\mathbf{G}_{\boldsymbol{\mu}}$ has learnt the distribution  $\mathbf{P}_{\boldsymbol{\mu'}}$ and hence satisfies the following equation:

\begin{equation}
\begin{aligned} 
 &\argmin_{\mathbf{G}_{\boldsymbol{\mu}}} ~~ \mathbb{E}_{\boldsymbol{\mu}\sim \mathbf{G}} [\mathit{Q}(\mathbf{A},\mathbf{E}_{\boldsymbol{\mu}})]  \\
  & \text{s.t.}~~\{\boldsymbol{\mu}: \boldsymbol{\mu} \sim \mathbf{G}_{\boldsymbol{\mu}}\} =  \{\boldsymbol{\mu'}: \boldsymbol{\mu'} \sim \mathbf{P}_{\boldsymbol{\mu'}} \}
\label{sup:objective-original}
\end{aligned}
\end{equation}

This concludes our empirical proof for our BBGAN.
\clearpage
\section{Special Cases of Generic Adversarial attacks:}
In Table \ref{tbl:var-summary-sup}, we summarize the substitutions in the generic adversarial attack to get different special cases of adversarial attacks. In summary, the generic adversarial attack allows for static agents (like classifiers and detectors) and dynamic agents (like an autonomous agent acting in a dynamic environment). It also covers the direct attack case of pixel perturbation on images to attack classifiers, as well as semantic attacks that try to fool the agent in a more realistic scenario (\eg camera direction to fool a detector). The generic attack also allows for a more flexible way to define the attack success based on an application-specific threshold $\epsilon$ and the agent score $\mathit{Q}$.
In the following we provide detailed explanation and mathematical meaning of the substitutions.

\begin{table*}[!t]
\small
\tabcolsep=0.09cm
\centering
 \begin{tabular}{||c||  c| c| c|| } 
 \hline
  \multicolumn{1}{||c||}{ } & \multicolumn{3}{|c||}{Substitutions of Special Cases of Adversarial Attacks }\\ 
  \hline  
 Generic Attack Variables& \specialcell{ Pixel Adversarial Attack\\ on Image Classifiers} & \specialcell{Semantic Adversarial Attack\\ on Object Detectors} & \specialcell{ Semantic Adversarial Attack\\ on Autonomous Agents.} \\
  \hline \hline
Agent $\mathbf{A}$	& \specialcell{K-class classifier \\$\mathbf{C}:[0,1]^{n} \rightarrow [l_{1}, l_{2}, ... ,l_{K}]$ \\ $l_{j}$ : the softmax value for class $j$ } &  \specialcell{K-class object detector \\ $\mathbf{F}:[0,1]^{n} \rightarrow  (\mathbb{R}^{N\times K},\mathbb{R}^{N\times 4})$ \\ $N$ : number of detected objects }	& \specialcell{self-driving policy agent $\mathbf{A}$\\ \eg network to regress controls } \\ \hline
Parameters $\boldsymbol{\mu}$ & \specialcell{ the pixels noise \\ added on attacked image}  & \specialcell{ some semantic parameters \\ describing the scene \\ \eg camera pose, object , light}  & \specialcell{ some semantic parameters \\ involved in the simulation  \\ \eg road shape , weather , camera}\\ \hline
Parameters Size $d$ & \specialcell{ $n$: the image dimension \\ $n = h\times w \times c$}  &  \specialcell{number of semantic parameters\\ parameterizing  $\mathbf{E}_{\boldsymbol{\mu}}$} &   \specialcell{number of semantic parameters\\ parameterizing $\mathbf{E}_{\boldsymbol{\mu}}$} \\ \hline
Environment $\mathbf{E}_{\boldsymbol{\mu}}$ & \specialcell{ dataset $\Phi$ containing all images\\ and their true class label \\ $\Phi = \{(\mathbf{x}_{i}, y_{i} )\}_{i=1}^{|\Phi|}$} &  \specialcell{ dataset $\Phi$ containing all images\\ and their true class label} & \specialcell{ simulation environment \\ partially described by $\boldsymbol{\mu}$\\ that $\mathbf{A}$ navigates in to reach}\\ \hline
Parameters Range $[\boldsymbol{\mu}_{\text{min}},\boldsymbol{\mu}_{\text{max}}]$ & \specialcell{$[-\mathbf{x}_{\text{min}},1-\mathbf{x}_{\text{max}}]$ \\ $\mathbf{x}_{\text{min}},\mathbf{x}_{\text{max}}$ are the min and max\\ of each pixel value in the image $\mathbf{x}$}
&  $[-1,1]^{d}$ & $[-1,1]^{d}$\\ \hline
Environment States $\mathbf{s}_{t}$ & \specialcell{static environment\\ $\mathbf{s}_{t} = \Phi$ } &  \specialcell{static environment\\ $\mathbf{s}_{t} = \Phi$ } & \specialcell{the state describing the simulation\\ at time step $t$ \\ \eg the new position of car }\\ \hline
Observation $\mathbf{o}_{t}$ &\specialcell{the attacked image \\after adding fooling noise} & \specialcell{the rendered image \\using the scene parameters $\boldsymbol{\mu}$ }  & \specialcell{the sequence of rendered images \\the $\mathbf{A}$ observe\\ during the simulation episode }\\ \hline
Episode Size $T$ & 1 &  1 & \specialcell{ $t_{\text{stop}} \in \{1,2,3,...T_{\text{max}}\}$\\ $T_{\text{max}}$ : max allowed time\\ $t_{\text{stop}}$ : step when success  reached  }\\ \hline
Time step $t$ & 1 &  1 & \specialcell{ time step where agent $\mathbf{A}$ has to \\decide on action $\mathbf{a}_{t}$ \\given what it has observed $\mathbf{o}_{t}$ }\\ \hline
Fooling Threshold $\epsilon$ & 0  & \specialcell{fixed small fraction of 1 \\ \eg = 0.3}  & \specialcell{fixed small fraction of 1 \\ \eg = 0.6}\\ \hline
Observation $\mathbf{o}_{t}$ &\specialcell{attacked image after added noise\\= $\mathbf{x}_{i} + \boldsymbol{\mu}$,  where  $\mathbf{x} ~,~ \boldsymbol{\mu} \in \mathbb{R}^{n}$ } & \specialcell{the rendered image \\using the scene parameters $\boldsymbol{\mu}$ }  & \specialcell{sequence of rendered images $\mathbf{A}$\\observes during the simulation episode}\\ \hline
Agent Actions $\mathbf{a}_{t}(\mathbf{o}_{t})$ & \specialcell{predicted softmax vector \\ of attacked image}  &  \specialcell{predicted confidence of \\ the true class label} & \specialcell{steering command to move\\ the car/UAV in the next step} \\ \hline
Reward Functions $\mathit{R}(\mathbf{s}_{t},\mathbf{a}_{t})$ & \specialcell{the difference between \\true and predicted softmax} &  \specialcell{predicted confidence of \\ the true class label} &\specialcell{1 : if success state reached \\0: if success not reached}\\ \hline
Score $\mathit{Q}(\mathbf{A},\mathbf{E}_{\boldsymbol{\mu}})$ & \specialcell{the difference between \\true and predicted softmax} &    \specialcell{predicted confidence of \\ the true class label} & \specialcell{the average sum of rewards \\ over five different episodes}  \\  \hline
 \hline 
 \end{tabular}
\caption{Special Cases of Generic Adversarial attacks: summarizing all the variable substitutions to get common adversarial attacks.}
\label{tbl:var-summary-sup}
\end{table*}

\subsection{Pixel Adversarial Attack on Image Classifiers}
Probably the most popular adversarial attack in the literature is a pixel-level perturbation to fool an image classifier. This attack can be thought of as a special case of our general formulation. In this case, the agent $\mathbf{A}$ is a classifier $\mathbf{C}: [0,1]^{n} \rightarrow [l_{1}, l_{2}, ... ,l_{K}]$ and the environment $\mathbf{E}_{\boldsymbol{\mu}}$ is a dataset containing the set $\Phi$ of all images in the classification dataset along with their respective ground truth labels, \ie  $\{(\mathbf{x}_{i}, y_{i} )\}_{i=1}^{|\Phi|}$ and $y_i\in\{1,\ldots,K\}$. The softmax value of the true class is given by $l_{y_i} = \max\mathbf{C}(\mathbf{x}_{i}) $. Parameter $\boldsymbol{\mu}$  defines the fooling noise to be added to the images (\ie $d=n$). The observation is simply an image from $\Phi$ with noise added:  $\mathbf{o_t} = \mathbf{x}_{i} + \boldsymbol{\mu} $ for some $i \in \{1,2,..,|\Phi|\}$. In classification, the environment is static with $T = 1$. 
To ensure the resulting image is in the admissible range, the noise added $\boldsymbol{\mu}$ should fall in the range $[-\mathbf{x}_{i,\text{min}},1-\mathbf{x}_{i,\text{max}}]$, where $\mathbf{x}_{i,\text{min}},\mathbf{x}_{i,\text{max}}$ are the min and max pixel value for the image $\mathbf{x}_{i}$.  
The sole action $\mathbf{a}_{1} $ is simply the softmax score of the highest scoring class label predicted by $\mathbf{C}$ that is not the true class $y_i$. Formally, $\mathbf{a}_{1} = \max_{j\neq y_i}\mathbf{C}(\mathbf{x}_{i}+\boldsymbol{\mu}) = l_{j}$. 
The reward function is $\mathit{R}(\mathbf{s}_{1},\mathbf{a}_{1}) = \mathit{Q}(\mathbf{C},\Phi) = \max (l_{y_i} - l_{j} ,0)$. Here, $\epsilon = 0 $ for the classifier fooling to occur , which means fooling occurs if $ l_{y_i} - l_{j}  \le 0 $. Using these substitutions in the hard constraint in \eqLabel{\ref{sup:fool-distribution}} translates to the following constraints on the perturbed image.

\begin{equation} 
\begin{aligned}
 l_{y_i} ~\le~ l_{j} ~,~ \mathbf{x}_{i} + \boldsymbol{\mu} \in [0,1]^{n}
\end{aligned}
\label{eq:constraint-transition}
\end{equation}
For a single image attack, we can rewrite \eqLabel{\ref{eq:constraint-transition}} as follows:
\begin{equation} 
\begin{aligned}
 \max\mathbf{C}(\mathbf{x}) ~\le~ \max_{j\neq y}\mathbf{C}(\mathbf{x'}) ~,~ \mathbf{x'} \in [0,1]^{n}
\end{aligned}
\label{eq:constraint-transition2}
\end{equation}
We observe that the constraints in \eqLabel{\ref{eq:constraint-transition2}} become the following constraints of the original adversarial pixel attack formulation on a classifier $\mathbf{C}$.

\begin{equation}
\begin{aligned} 
\resizebox{0.90\hsize}{!}{
 $\underset{\mathbf{x'}\in [0,1]^{n}}{\min}  \mathit{d}(\mathbf{x},\mathbf{x'}) ~~\text{s.t.}~~   \argmax~ \mathbf{C}(\mathbf{x}) \ne \argmax~ \mathbf{C}(\mathbf{x'})$}
\label{eq:classifier-attack}
\end{aligned}
\end{equation}

\subsection{Semantic Adversarial Attack on Object Detectors} \label{detector-attack}
Extending adversarial attacks from classifiers to object detectors is straight-forward. We follow previous work  \cite{shapeshifter} in defining the object detector as a function $\mathbf{F} : [0,1]^{n} \rightarrow (\mathbb{R}^{N\times K},\mathbb{R}^{N\times 4})$, which takes an $n$-dimensional image as input and outputs $N$ detected objects. Each detected object has a probability distribution over $K$ class labels and a 4-dimensional bounding box for the detected object. We take the top $J$ proposals according to their confidence and discard the others. Analyzing the detector in our general setup is similar to the classifier case. The  environment $\mathbf{E}_{\boldsymbol{\mu}}$ is static (\ie $T=1$), and it contains all images with ground truth detections. For simplicity, we consider one object of interest per image (indexed by $i$). 
The observation in this case is a rendered image of an instance of object class $i$, where the environment parameter $\boldsymbol{\mu}$ determines the 3D scene and how the image is rendered (\eg the camera position/viewpoint, lighting directions, textures, \etc.). Here, the observation is defined as the rendering function $\mathbf{o}_1: [\boldsymbol{\mu}_{\text{min}},\boldsymbol{\mu}_{\text{max}}]^{d} \rightarrow \mathbb{R}^{n}$. We use Blender \cite{blender} to render the 3D scene containing the object and to determine its ground truth bounding box location in the rendered image. 
The action $\mathbf{a}_{1}$ by the agent/detector is simply the highest confidence score $l_{i}$ corresponding to  class $i$ from the top $J$ detected boxes in $\mathbf{o}_1$. The final score of $\mathbf{F}$ is $\mathit{Q}(\mathbf{F},\mathbf{E}_{\boldsymbol{\mu}}) = l_{i} $. The attack on $\mathbf{F}$ is considered successful, if $l_{i} \leq \epsilon$.

\subsection{Semantic Adversarial Attack on Autonomous Agents}
The semantic adversarial attack of an autonomous agent can also be represented in the general formulation of Algorithm 1 in the paper. Here, $\mathbf{A}$ corresponds to the navigation policy, which interacts with a parametrized environment $\mathbf{E}_{\boldsymbol{\mu}}$. The environment parameter $\boldsymbol{\mu} \in \mathbb{R}^{d}$ comprises $d$ variables describing the weather/road conditions, camera pose, environment layout \etc. 
In this case, an observation $\mathbf{o}_t$ is an image as seen from the camera view of the agent at time $t$. The action $\mathbf{a}_t$ produced by the navigation policy is the set of control commands (\eg gas and steering for a car or throttle, pitch, roll and yaw for a UAV). The reward function $\mathit{R}(\mathbf{s}_{t},\mathbf{a}_{t})$ measures if the agent successfully completes its task (\eg $1$ if it safely reaches the target position at time $t$ and $0$ otherwise). 
The episode ends when either the agent completes its task or the maximum number of iterations $T$ is exceeded. Since the reward is binary, the $\mathit{Q}$ score is the average reward over a certain number of runs (five in our case). This leads to a fractional score $ 0 \le \mathit{Q} \le 1$.
\clearpage
\clearpage
\section{Boosting Strategy for BBGAN } \label{sec:boosting}
\subsection{Intuition for Boosting}
Inspired by the classical Adaboost meta-algorithm \cite{adaboost}, we use a boosting strategy to improve the performance of our BBGAN trained in Section 5 of our paper with results reported in Table 1. The boosting strategy of BBGAN is simply utilizing the information learned from one BBGAN by another BBGAN in a sequential manner. The intuition is that the main computational burden in training the BBGAN is not the GAN training but computing the agent $\mathbf{A}$ episodes (which can take multiple hours per episode in the case of the self-driving experiments).

\subsection{Description of Boosting for BBGANs} \label{sec:boosting-desc}
We propose to utilize the generator to generate samples that can be used by the next BBGAN. We start by creating the set $\Omega_{0} $ of the first stage adversary $\mathbf{G}_{0}$. We then simply add the generated parameters $\boldsymbol{\mu}$ along with their computed scores $\mathit{Q}$ to the training data of the next stage BBGAN (\ie BBGAN-1). We start the next stage by inducing a new induced set $S_{\boldsymbol{\mu'}}^{1}$ (that may include part or all the previous stage induced set $S_{\boldsymbol{\mu'}}^{0}$). However, the aim is to put more emphasis on samples that were generated in the previous stage. Hence, the inducer in the next stage can just randomly sample N points, compute their $\mathit{Q}$ scores and add $\beta * N$ generated samples from BBGAN-0 to the $N$ random samples. The entire set is then sorted based on the $\mathit{Q}$ scores, where the lowest-scoring $s_{1}$ points that satisfy \eqLabel{\ref{sup:fool-distribution}} are picked as the induced set $S_{\boldsymbol{\mu'}}^{1},~ s_{1} = \left|S_{\boldsymbol{\mu'}}^{1}\right|$. The BBGAN-1 is then trained according to \eqLabel{\ref{sup:BBGAN}}. Here $\beta$ is  the boosting rate of our boosting strategy which dictates how much emphasis is put on the previous stage (exploitation ratio) when training the next stage. The whole boosting strategy can be repeated more than once. The global set $\Omega$ of all $N$ sampled points and the update rule from one stage to another is described by the following two equations:

\begin{equation}
\begin{aligned} 
\Omega_{0} ~ =  ~\{\boldsymbol{\mu}_{j} \sim \text{Uniform}(\boldsymbol{\mu}_{\text{min}},\boldsymbol{\mu}_{\text{max}}) \}_{j=1}^{N}
\label{sup:omega}
\end{aligned}
\end{equation}

\begin{equation}
\begin{aligned} 
\Omega_{k} ~ = \Omega_{k-1} \cup ~\{\boldsymbol{\mu}_{j} \sim \mathbf{G}_{k-1} \}_{j=1}^{\lfloor \beta  N\rfloor}
\label{sup:boost-omega}
\end{aligned}
\end{equation}

These global sets $\Omega_{k}$ constitute the basis from which the inducer produces the induced sets $S_{\boldsymbol{\mu'}}^{k}$. The adversary $\mathbf{G}_{k}$ of boosting stage $k$ uses this induced set when training according to the BBGAN objective in \eqLabel{\ref{sup:BBGAN}}. Algorithm \ref{alg: boost} summarizes the boosting meta-algorithm for BBGAN.

\begin{algorithm}[!t] 
\caption{Boosting Strategy for BBGAN}\label{alg: boost}
\small
\SetAlgoLined
  \textbf{Requires: } environment $ \mathit{E_{\boldsymbol{\mu}}}$, Agent $\mathbf{A}$ number of boasting stages $K$, boosting rate $\beta$, initial training size $N$ \\
      Sample $N$ points to form  $\Omega_{0}$ like in \eqLabel{\ref{sup:omega}} \\
      induce $ S_{\boldsymbol{\mu'}}^{0}$ from $\Omega_{0}$ \\
      learn adversary $\mathbf{G}_{0}$ according to \eqLabel{\ref{sup:BBGAN}}\\
  \For{$i \leftarrow 1$ \KwTo $K$}{
  update boosted training set $\Omega_{i}$ from $\Omega_{i-1}$ as in \eqLabel{\ref{sup:boost-omega}}  \\
  obtain $ S_{\boldsymbol{\mu'}}^{i}$ from $\Omega_{i}$ \\
  train adversary $\mathbf{G}_{i}$ as in \eqLabel{\ref{sup:BBGAN}}
}

\textbf{Returns: } last adversary $\mathbf{G}_{K}$
\end{algorithm}

\subsection{Empirical proof for BBGAN Boosting} \label{sec:boosting-proof}
Here we want to show the effectiveness of boosting (Algorithm \ref{alg: boost}) on improving the performance of BBGAN from one stage to another. Explicitly, we want to show that the following statement holds, under some conditions.

\begin{equation} 
\begin{aligned}
\mathbb{E}_{\boldsymbol{\mu}_{k}\sim \mathbf{G}_{k}} [\mathit{Q}(\mathbf{A},\mathbf{E}_{\boldsymbol{\mu}_{k}})] \le \mathbb{E}_{\boldsymbol{\mu}_{k-1}\sim \mathbf{G}_{k-1}} [\mathit{Q}(\mathbf{A},\mathbf{E}_{\boldsymbol{\mu}_{k-1}})]
\end{aligned}
\label{sup:boosting-proof}
\end{equation}
This statement says that the expected score $\mathit{Q}$ of the sampled parameters $\boldsymbol{\mu}_{k}$ from the adversary $\mathbf{G}_{k}$ of stage $(k)$ BBGAN is bounded above by the score of the previous stage , which indicates iterative improvement of the fooling adversary $\mathbf{G}_{k}$ by lowering the score of the agent $\mathbf{A}$ and hence achieving a better objective at the following realxation of  \eqLabel{\ref{sup:objective}}.
\begin{equation}
\begin{aligned} 
 &\argmin_{\mathbf{G}} ~~ \mathbb{E}_{\boldsymbol{\mu}\sim \mathbf{G}} [\mathit{Q}(\mathbf{A},\mathbf{E}_{\boldsymbol{\mu}})]  \\
  & \text{s.t.}~~\{\boldsymbol{\mu}: \boldsymbol{\mu} \sim \mathbf{G}\} \subset \{\boldsymbol{\mu'}: \boldsymbol{\mu'} \sim \mathbf{P}_{\boldsymbol{\mu'}} \}
\label{sup:relax}
\end{aligned}
\end{equation}
\mysection{Proof of \eqLabel{\ref{sup:boosting-proof}} }\\
Let's start by sampling random $N$ points as our initial $\Omega_{0}$ set as in \eqLabel{\ref{sup:omega}} and then learn BBGAN of the first stage and continue boosting as in Algorithm \ref{alg: boost}. Assume the following assumption holds,    

\begin{equation} 
\begin{aligned}
\left|S_{\boldsymbol{\mu}}^{k}\right| =  \lfloor \beta  N\rfloor ~~,~ \forall~ k \in \{1,2,3,...\} 
\end{aligned}
\label{sup:boosting-assum}
\end{equation}
then by comparing the average score $\mathit{Q}$ of the entire global set $\Omega_{k}$ at stage $k$ (denoted simply as $\mathit{Q}(\Omega_{k}))$) with the average score of the added boosting samples from the previous stage $~\{\boldsymbol{\mu}_{j} \sim \mathbf{G}_{k-1} \}_{j=1}^{\lfloor \beta  N\rfloor}$ as in \eqLabel{\ref{sup:boost-omega}}  (denoted simply as $\mathit{Q}(\mathbf{G}_{k-1})$) , two possibilities emerge:

\vspace{3pt} \mysection{1. Exploration possibility: $  \mathit{Q}(\Omega_{k})) \le \mathit{Q}(\mathbf{G}_{k-1})  $  } \\
This possibility indicates that there is at least one \textit{new} sample in the global set $\Omega_{k}$ that are not inherited from the previous stage adversary $\mathbf{G}_{k-1}$, which is strictly  better then $\mathbf{G}_{k-1}$ samples with strictly lower $\mathit{Q}$ score. If the assumption in \eqLabel{\ref{sup:boosting-assum}}holds, then the induced set $S_{\boldsymbol{\mu}}^{k}$ will include at least one new parameter that is not inherited from previous stage and hence the average score of the induced set will be less than that of the generated by previous stage.  
\begin{equation} 
\begin{aligned}
\mathit{Q}(S_{\boldsymbol{\mu}}^{k}) < \mathit{Q}(\mathbf{G}_{k-1})
\end{aligned}
\label{sup:boosting-phenom}
\end{equation}
However since the BBGAN of stage $k$ uses the induced set $S_{\boldsymbol{\mu}}^{k}$ for training , we expect the samples to be correlated: $\mathbf{G}_{k} \sim S_{\boldsymbol{\mu}}^{k}$, and the scores to be similar as follows: 

\begin{equation} 
\begin{aligned}
\mathbb{E}_{\boldsymbol{\mu}_{k}\sim \mathbf{G}_{k}} [\mathit{Q}(\mathbf{A},\mathbf{E}_{\boldsymbol{\mu}_{k}})] = \mathit{Q}(S_{\boldsymbol{\mu}}^{k}) 
\end{aligned}
\label{sup:boosting-corr}
\end{equation}
Substituting \eqLabel{\ref{sup:boosting-corr}} in \eqLabel{\ref{sup:boosting-phenom}} results in the inequality

which makes the less strict inequality \eqLabel{\ref{sup:boosting-proof}}holds.

\vspace{3pt} \mysection{2. Exploitation possibility: $  \mathit{Q}(\Omega_{k})) > \mathit{Q}(\mathbf{G}_{k-1})  $  } \\
In this scenario, we don't know for sure whether there is a new sample in $\Omega_{k}$ that is better than the inherited samples, but in the worst case scenario we will get no new sample with lower score. In either case, the assumption in \eqLabel{\ref{sup:boosting-assum}} ensures that the new induced set $S_{\boldsymbol{\mu}}^{k}$ is exactly the inherited samples from $\mathbf{G}_{k-1}$ and the following holds.
\begin{equation} 
\begin{aligned}
\mathit{Q}(S_{\boldsymbol{\mu}}^{k}) \le \mathit{Q}(\mathbf{G}_{k-1})
\end{aligned}
\label{sup:boosting-phenom2}
\end{equation}
Using the same argument as in \eqLabel{\ref{sup:boosting-corr}}, we deduce that in this exploitation scenario \eqLabel{\ref{sup:boosting-proof}} is still satisfied. Hence, we prove that \eqLabel{\ref{sup:boosting-proof}} holds given the assumption in \eqLabel{\ref{sup:boosting-assum}}.

\subsection{Experimental Details for Boosting} \label{sec:boosting-exp}
We note that low $\beta$ values do not affect the training of our BBGAN since the induced set will generally be the same. Hence, we use $\beta = 0.5$, a high boosting rate. For practical reasons (computing 50\% of the training data per boosting stage is expensive) we just compute 10\% of the generated data and repeat it 5 times. This helps to stabilize the BBGAN training and forces it to focus more on samples that have low scores without having to evaluate the score function on 50\% of the training data. 

\clearpage

\section{Detailed Results}
Tables \ref{detector-table} and \ref{navigation-table} show the detailed results for all three applications.

\section{Nearest Neighbor Visualization} 
\begin{figure}[b]
\includegraphics[]{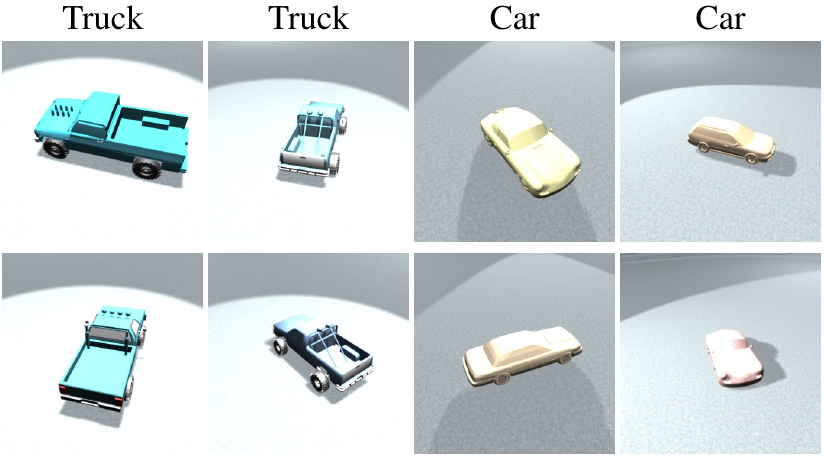}
\caption{\small \textbf{Nearest Neighbor from Training}: \emph{top}: generated fooling samples by our BBGAN for different classes. \emph{bottom}: the corresponding NN from the training. From left to right, $L_{2}$ norm differences in parameter space : (0.76, 0.45, 0.60, 0.50) ,$L_{2}$ norm differences in image space : (379, 218, 162, 164).}
\label{fig:nn1}
\end{figure}

\begin{figure}[t]
\includegraphics[]{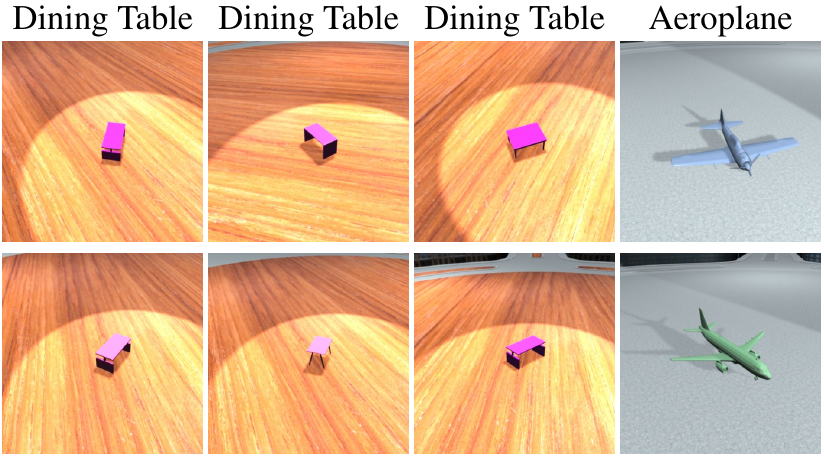}
\caption{\small \textbf{Nearest Neighbor from Training}: \emph{top}: generated fooling samples by our BBGAN for different classes. \emph{bottom}: the corresponding NN from the training. From left to right, $L_{2}$ norm differences in parameter space : (0.60, 0.50, 0.87, 0.47) ,$L_{2}$ norm differences in image space : (124, 125, 215, 149).}
\label{fig:nn2}
\end{figure}

\begin{figure}[b]
\includegraphics[]{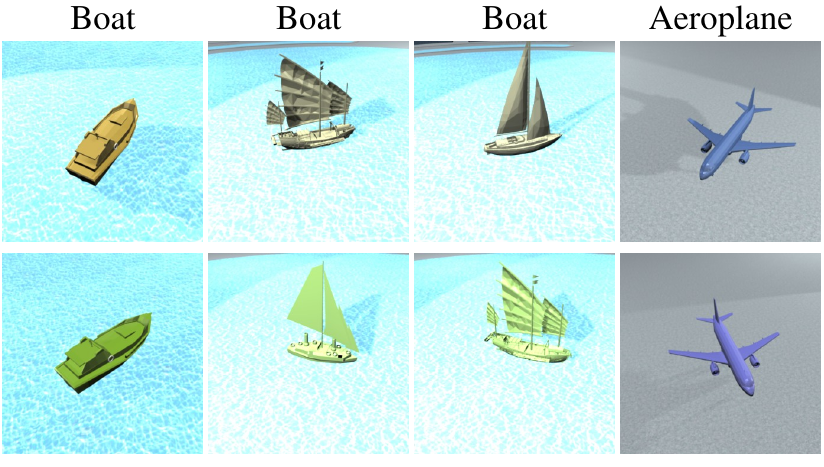}
\caption{\small \textbf{Nearest Neighbor from Training}: \emph{top}: generated fooling samples by our BBGAN for different classes. \emph{bottom}: the corresponding NN from the training. From left to right, $L_{2}$ norm differences in parameter space : (0.56, 0.44, 0.43, 0.81) ,$L_{2}$ norm differences in image space : (174, 197, 190, 99).}
\label{fig:nn3}
\end{figure}

In \figLabel\ref{fig:nn1}, \figLabel\ref{fig:nn2} and \figLabel\ref{fig:nn3} we visualize the NN in the parameter space for four different generated samples by our BBGAN. We see that the generated and the NN in training are different for the 4 samples with $L_{2}$ norm differences in the caption of each figure. All pixels and parameters range from -1 to 1. This shows that our BBGAN can generate novel examples that fool the trained agent.

\begin{table*}[!h]
\small
\tabcolsep=0.08cm
\centering
 \begin{tabular}{||c| c c c  c c c c c c c c c c| c||} 
 \hline
\multicolumn{1}{||c}{} & \multicolumn{13}{c}{\text{Adversarial Attack Fooling Rate across Different Classes}} &  \multicolumn{1}{c||}{} \\
   & aeroplane & bench & bicycle & boat & bottle & bus & car & chair & diningtable & motorbike & train & truck & avg & $\mu_{std}$\\ 
 \hline\hline
 {Full Set} & 8.64\%        & 35.2\%        & 14.6\%        & 33.4\%        & 22.5\%    & 53.1\%        & 39.8\%        & 44.1\%        & 46.1\%        & 32.5\%        &     58.1\%   &       56.8\% &  37.1 \% & 0.577 \\
 {Random:} & 11.3\%        & 42.7\%        & 18.6\%        & 41.8\%        & 28.4\%    & 65.7\%        & 49.9\%        & 55.3\%        & 56.4\%        & 40.3\%        &     72.8\%   &       70.8\% &  46.2\% & 0.584 \\
  {Multi-Class SVM} & 12.0\%      & 45.6\%      & 20.0\%    & 39.6\%    & 26.0\%      & 64.4\%      & 49.6\%  & 50.4\%       & 53.6\%      & 45.6\%      & 72.0\%      & 70.8\%    & 45.8\%  & 0.576 \\
 {GP Regression} & 13.6\%      & 15.6\%      & 17.6\%      & 41.2\%      & 31.6\%  & 71.6\%       & 51.6\%      & 48.0\%      & 56.0\%      & 43.6\%      & 69.2\%     & 83.6\%    & 45.26\%  & 0.492 \\
 \hline \hline
 {Gaussian} & 11.2\%       & 45.6\%       & 19.6\%       & 41.6\%       & 31.2\%   & 70.4\%       & 48.0\%       & 56.8\%       & 55.6\%       & 40.4\%       & 71.2\%      & 72.4\% &  47.0\%  & 0.548 \\
 {GMM10\%} & 14.8\%       & 45.2\%       & 26.0\%       & 42.8\%       & 34.0\%   & 67.2\%    & 53.2\%       & 56.4\%       & 54.8\%       & 48.4\%       & 70.4\%      & 75.2\%    & 49.0\%  &  0.567 \\
 {GMM50\%} & 12.0\%       & 44.0\%       & 16.4\%       & 46.4\%       & 33.2\%   & 66.4\%       & 51.6\%       & 53.2\%       & 58.4\%       & 46.8\%       & 73.6\%      & 72\% &  47.8\%   & 0.573 \\
 {Bayesian} & 9.2\%      &  42.0\%      & 48.0\%      & 68.8\%      & 32.4\%  & 91.6\%      & 42.0\%      & 75.6\%      & 58.4\%      & 52.0\%      & 77.2\%     & 75.6\%    & 56.1\% & 0.540 \\
 {BBGAN (ours)} & 13.2\%      & 91.6\%      & 44.0\%      & 90.0\%      & 54.4\%  & 91.6\%      & 81.6\%      & 93.2\%      & 99.2\%      & 45.2\%      & 99.2\%     & 90.8\%  & 74.5\%   & 0.119 \\ 
 {BBGAN (boost)} & 33\%      & 82.4\%      & 65.8\%      & 78.8\%      & 67.4\%  & 100\%      & 67.4\%      & 100\%      & 90.2\%      & 82.0\%      & 98.4\%     & 100\%   &  80.5\%      & 0.100 \\ [1ex] 

 \hline 
 \end{tabular}
\caption{Fooling rate of adversarial attacks on different classes of the augmented Pascal3D dataset. We sample 250 parameters after the training phase of each model and sample a shape from the intended class. We then render an image according to these parameters and run the YOLOV3 detector to obtain a confidence score of the intended class. If this score $\mathit{Q} \leq \epsilon = 0.3$, then we consider the attack successful. The fooling rate is then recorded for that model, while $\boldsymbol{\mu}_{\text{std}}$ (the mean of standard deviations of each parameter dimensions) is recorded for each model. We report the average over all classes. This metric represents how varied the samples from the attacking distribution are.}
\label{detector-table}
\end{table*}

\begin{table*}[!h]
\small
\begin{tabular}{||l|cccccccccccc||}
\hline
& \multicolumn{2}{c}{Straight} & \multicolumn{2}{c}{One Curve} & \multicolumn{2}{c}{Navigation} & \multicolumn{2}{c}{3 control points} & \multicolumn{2}{c}{4 control points} & \multicolumn{2}{c||}{5 control points}\\ 
& FR & $\mu_{std}$& FR & $\mu_{std}$& FR & $\mu_{std}$ & FR & $\mu_{std}$& FR & $\mu_{std}$& FR & $\mu_{std}$ \\ \hline
Full Set          & 10.6\%   & 0.198   & 19.5\%   & 0.596  & 46.3\%   & 0.604 & 17.0\% & 0.607 & 23.5\% & 0.544 & 15.8\% & 0.578 \\
Random                  & 8.0\%    & 0.194   & 18.0\%   & 0.623  & 48.0\%   & 0.572 & 22.0\% & 0.602 & 30.0\% & 0.550 & 16.0\% & 0.552 \\
Multi-Class SVM  & 96.0\%   & 0.089   & 100\%    & 0.311  & 100\%    & 0.517 & 24.0\% & 0.595 & 30.0\% & 0.510 & 14.0\%  & 0.980 \\
GP Regression    & 100\%    & 0.014   & 100\%    & 0.268  & 100\%    & 0.700 & 74.0\% & 0.486 & 94.0\% & 0.492 & 44.0\%  & 0.486 \\ 
Gaussian                & 54.0\%   & 0.087   & 30.0\%   & 0.528  & 64.0\%   & 0.439 & 49.3\% & 0.573 & 56.0\% & 0.448 & 28.7\% & 0.568 \\
GMM10\%                 & 90.0\%   & 0.131   & 72.0\%   & 0.541  & 98.0\%   & 0.571 & 57.0\% & 0.589 & 63.0\% & 0.460 & 33.0\% & 0.558 \\
GMM50\%                 & 92.0\%   & 0.122   & 68.0\%   & 0.556  & 100\%    & 0.559 & 54.0\% & 0.571 & 60.0\% & 0.478 & 40.0\% & 0.543 \\
\hline
BBGAN (ours)            & 100\%    & 0.048   & 98.0\%   & 0.104  & 98.0\%         & 0.137     & 42.0\% & 0.161 & 94.0\%  & 0.134 & 86.0\%  & 0.202 \\ 
BBGAN (boost)    & 100\%  & 0.014       & 100\%  & 0.001       & 100\%  & 0.058         & 86.0\%  &   0.084       & 98.0\%  & 0.030      & 92.0\%  & 0.003       \\
\hline
\end{tabular}
\caption{Autonomous Driving (CARLA) and UAV Racing Track Generation (Sim4CV). Each method produces 50 samples and we show the fooling rate (FR) and the mean of the standard deviation per parameter. We set the fooling threshold to 0.6 and 0.7 for autonomous driving and racing track generation respectively.}
\label{navigation-table}
\end{table*}

\section{Qualitative Examples }

\begin{figure*}
\centering
\includegraphics[width=0.82\textwidth]{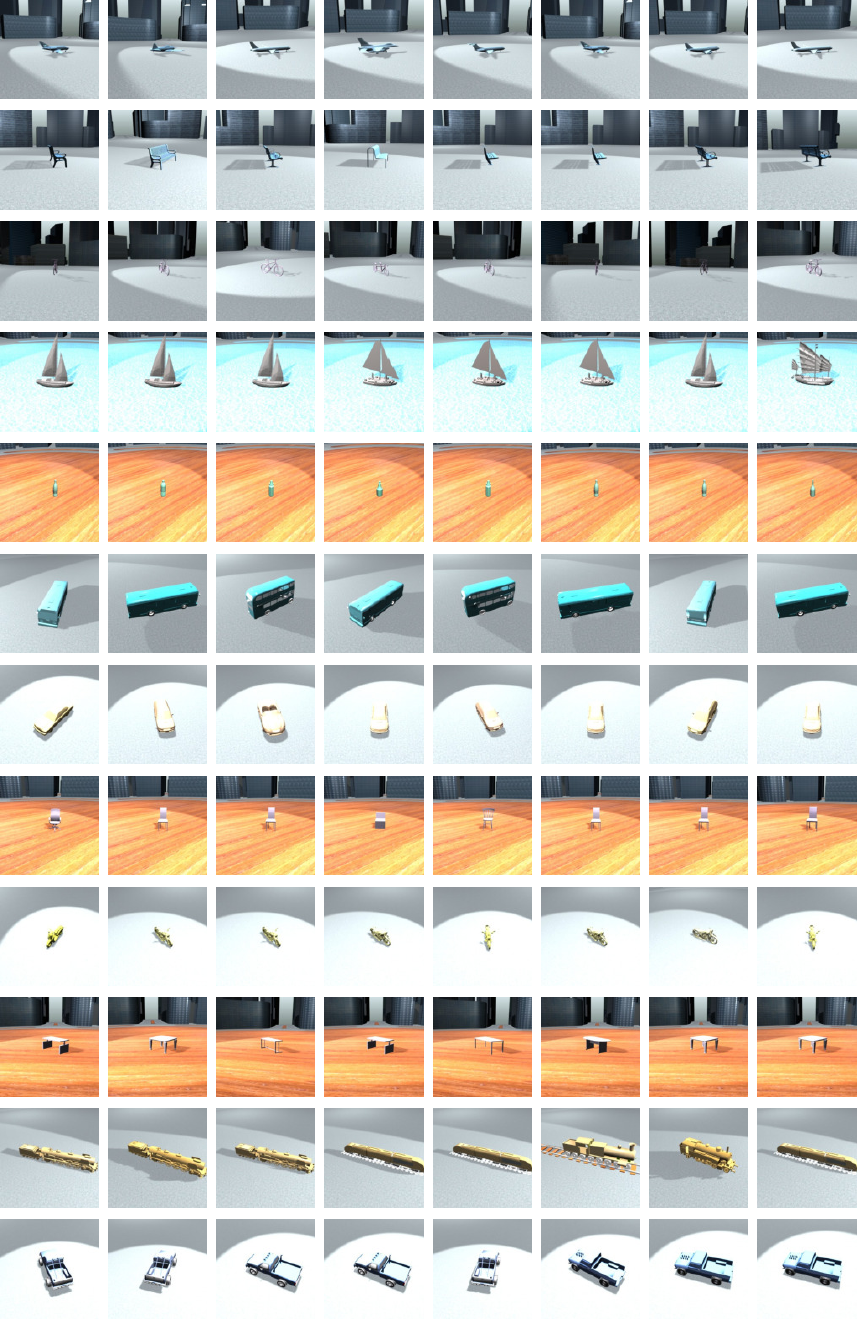}
\captionof{figure}{\textbf{BBGAN Qualitative Examples in Object Detection} - Some sample images for each class that were rendered according to parameters generated by BBGAN which fooled the object detector. }
\label{fig:qualitive_results_yolo}
\end{figure*}

\figLabel \ref{fig:qualitive_results_yolo} shows some qualitative examples for each of the 12 object classes. These images were rendered according to parameters generated by BBGAN which fooled the detector.  

\begin{figure*}
\centering
\includegraphics[]{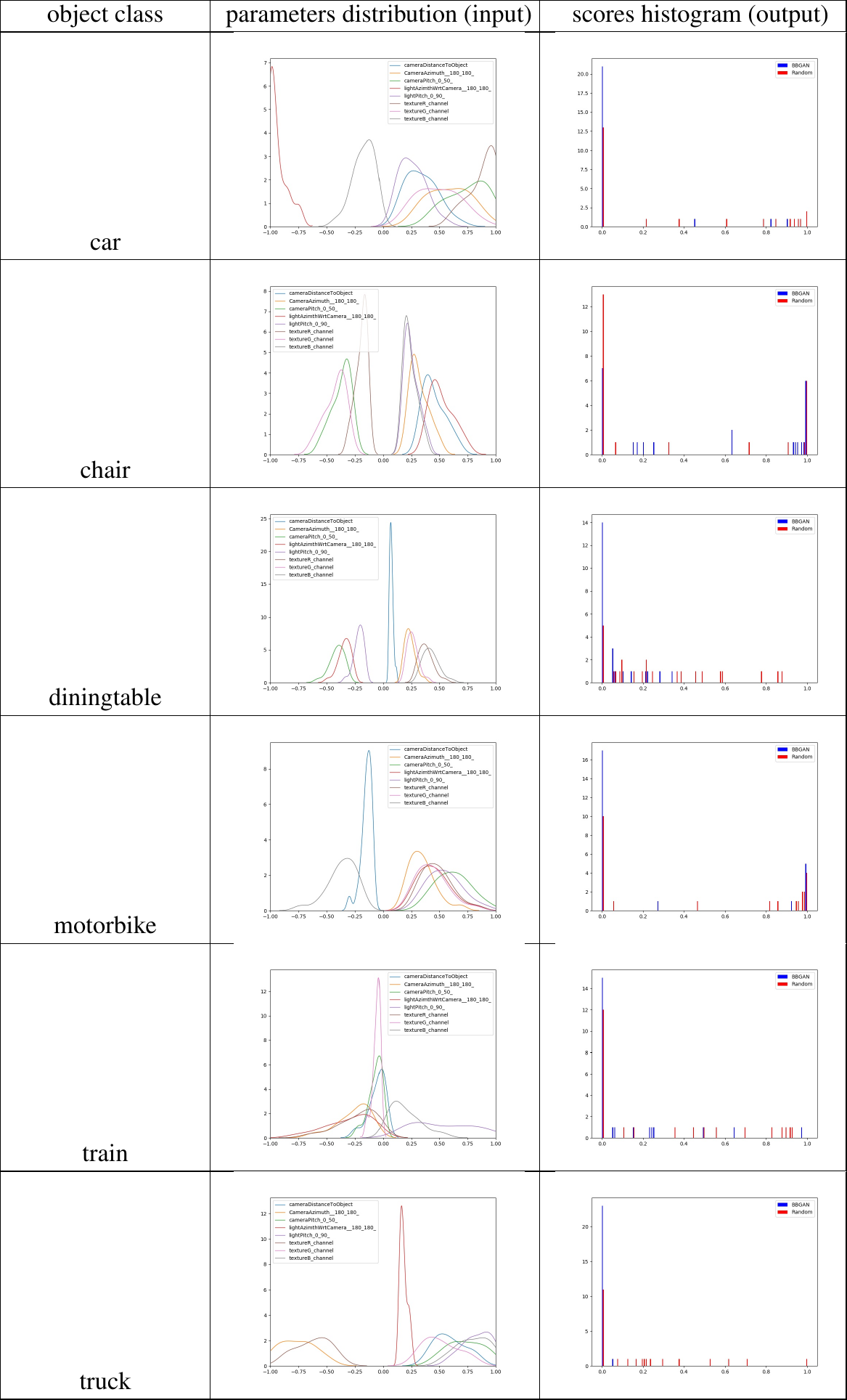}
\caption{\small \textbf{BBGAN Distribution Visualization 1}: visualizing the input parameters marginal distributions (the range is normalized from -1 to 1). Also, the Agent scores histogram for these parameters vs random parameters scores histogram are shown in the right column.}
\label{fig:dist-yolo1}
\end{figure*}

\begin{figure*}
\centering
\includegraphics[]{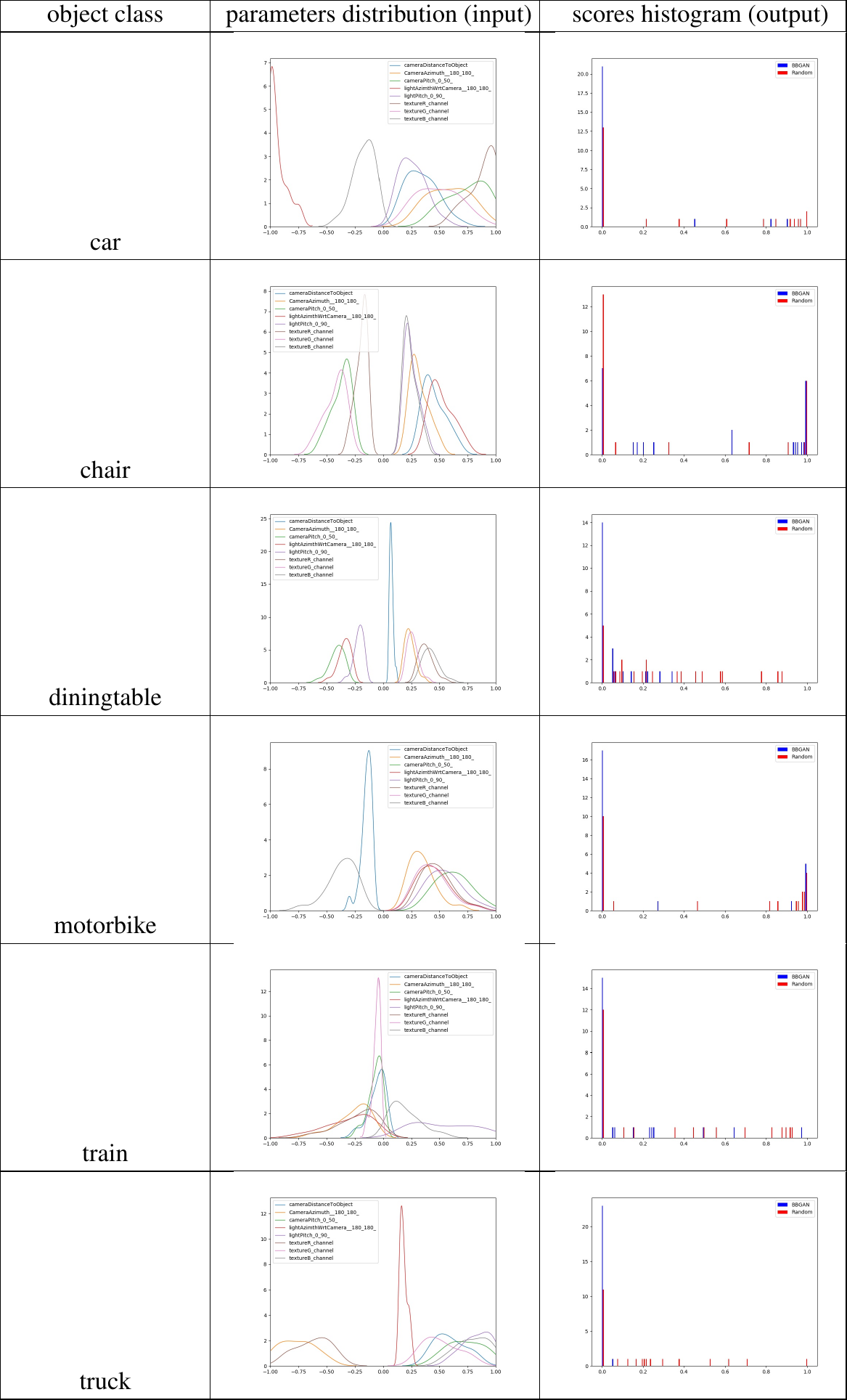}
\caption{\small \textbf{BBGAN Distribution Visualization 2}: visualizing the input parameters marginal distributions (the range is normalized from -1 to 1). Also, a histogram of agent scores for generated parameters and a histogram of scores for random parameters are shown in the right column.}
\label{fig:dist-yolo2}
\end{figure*}

\section{Qualitative Comparison} \label{sup:comparison}
\begin{figure*}
\centering
\includegraphics[]{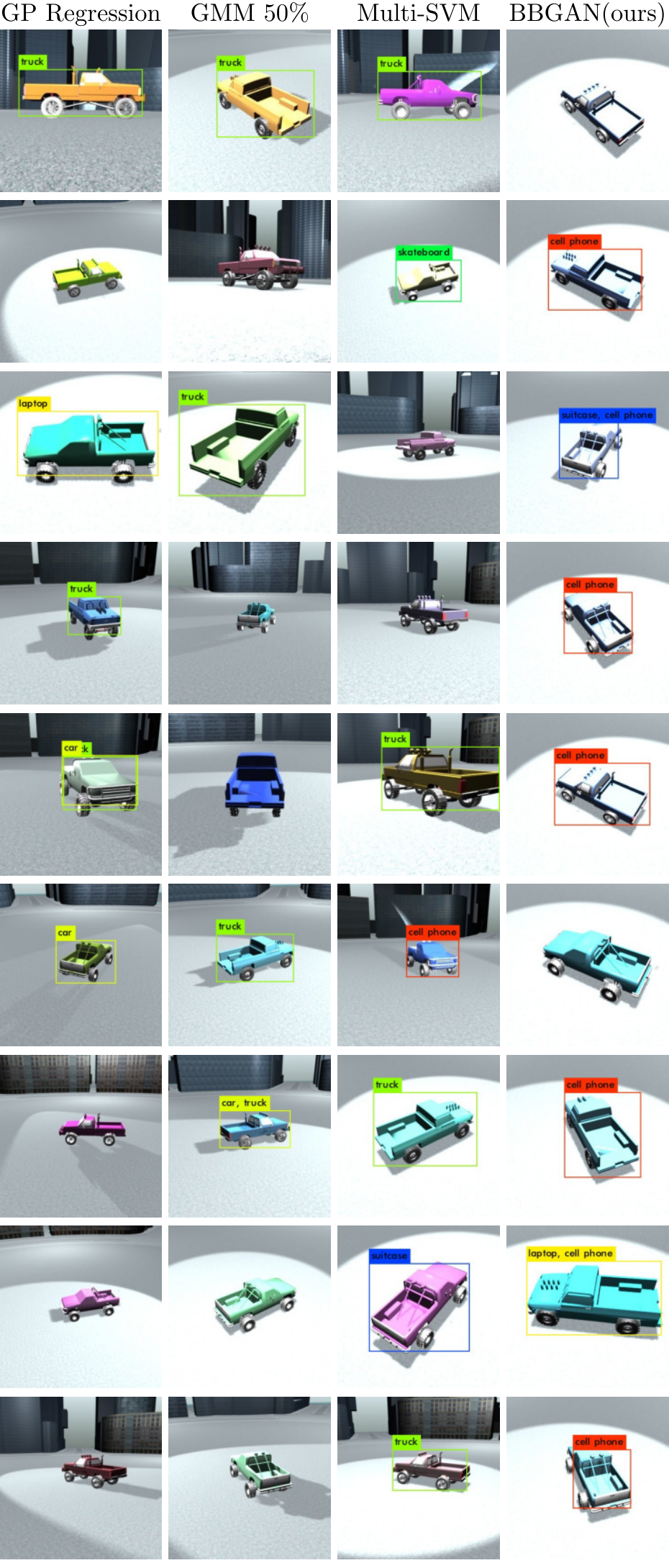}
\caption{\small \textbf{Qualitative Comparison for YOLOV3 Experiments}: Comparing the distribution of the best baselines with the distribution learned by our BBGAN. The samples shown are from the truck class experiment.}
\label{fig:qual-comp-yolo}
\end{figure*}

\figLabel{\ref{fig:qual-comp-yolo}} shows a qualitative comparison between samples of the BBGAN distribution and different baselines distributions in the YOLOYV3 attacks experiments. 
\subsection{Baselines}

\mysection{BBGAN}\label{training}
To learn the fooling distribution $\mathbf{P}_{\boldsymbol{\mu'}}$, we train the BBGAN as described in Methodology \secLabel. For this, we use a vanilla GAN model \cite{GAN}. We use a simple MLP with 2 layers for the Generator $\mathbf{G}$ and Discriminator $\mathbf{D}$. We train the GAN following convention, but since we do not have access to the true distribution that we want to learn (\ie real samples), we \emph{induce} the set by randomly sampling $N$ parameter vector samples $\boldsymbol{\mu}$, and then picking the $s$ worst  among them (according to $Q$ score). For object detection, we use $N=20000$ image renderings for each class (a total of 240K images), as described in the Application \secLabel. 
Due to the computational cost, our dataset for the autonomous navigation tasks comprises only $N=1000$ samples. For instance, to compute one data point in autonomous driving, we need to run a complete episode that requires 15 minutes. The induced set size $s$ is always fixed to be 100. 

\mysection{Random} We uniformly sample random parameters $\boldsymbol{\mu}$.

\mysection{Gaussian Mixture Model (GMM)} We fit a full covariance GMM of varying Gaussian components to estimate the distribution of the samples in the induced set $S_{\boldsymbol{\mu'}}$. The variants are denoted as Gaussian (one component), GMM10\% and GMM50\% (number of components as percentage of the samples in the induced set).

\mysection{Bayesian} We use the Expected Improvement (EI) Bayesian Optimization algorithm \cite{expected-improvment} to minimize the score $\mathit{Q}$ for the agent. The optimizer runs for 10$^4$ steps and it tends to gradually sample more around the global minimum of the function. So, we use the last $N=1000$ samples to generate the induced set $S_{\boldsymbol{\mu'}}$ and then learn a GMM on that with different Gaussian components. Finally, we sample $M$ parameter vectors from the GMMs and report results for the best model.

\mysection{Multi-Class SVM}
We bin the score $Q$ into 5 equally sized bins and train a multi-class SVM classifier on the complete set $\Omega$ to predict the correct bin.
We then randomly sample parameter vectors $\boldsymbol{\mu}$, classify them, and sort them by the predicted score. We pick $M$ samples with the lowest  $Q$ score.

\mysection{Gaussian Process Regression}
Similar to the SVM case, we train a Gaussian Process Regressor \cite{gp-robot} with an exponential kernel to regress $Q$ scores from the corresponding $\boldsymbol{\mu}$ parameters that generated the environment on the dataset $\Omega$. 

\clearpage 

\section{Training data} \label{training-data}
In the following we show some examples of the training data for each of the applications.
\subsection{Object Detection} \label{sec:application-detection}
Object detection is one of the core perception tasks commonly used in autonomous navigation. 
Its goal is to determine the bounding box and class label of objects in an image. You Only Look Once (YOLO) object detectors popularized a single-stage approach, in which the detector observes the entire image and regresses the boundaries of the bounding boxes and the classes directly \cite{yolo}. This trades off the accuracy of the detector for speed, making real-time object detection possible.

\mysection{Agent}
Based on its suitability for autonomous applications, we choose the very fast, state-of-the-art YOLOv3 object detector \cite{yolo3} as the agent in our SADA framework. It achieves a competitive mAP score on the MS-COCO detection benchmark and it can run in real-time \cite{coco}. 

\begin{figure*}
\centering
\includegraphics[width=\textwidth]{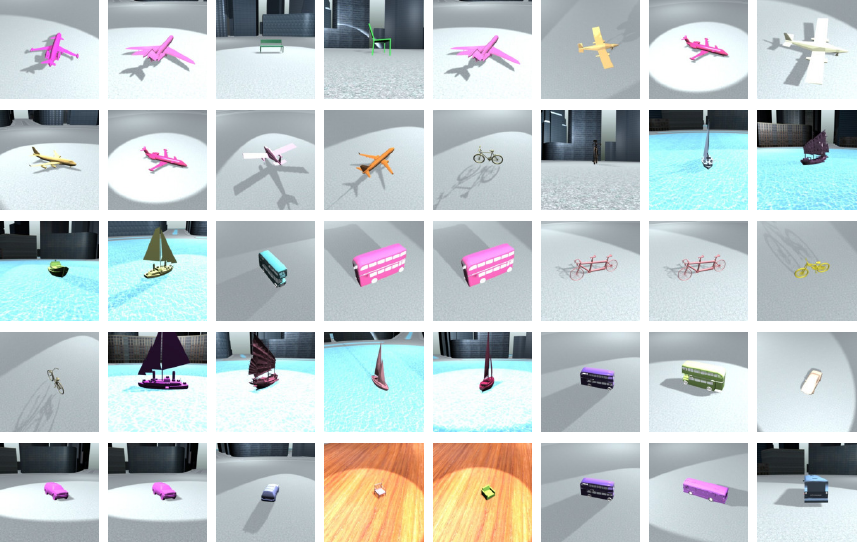}
\captionof{figure}{\textbf{Training Data for YOLOV3 Experiment } - Some sample images from the dataset used for object detection with YOLOV3. Note that in the actual dataset each object has a random color regardless of its class. For clarity we uniformly color each class in this figure.}
\label{fig:qualitive_dataset_yolo}
\end{figure*}

\mysection{Environment}
We use Blender open source software to construct a scene based on freely available 3D scenes and CAD models . The scene was picked to be an urban scene with an open area to allow for different rendering setups. The scene includes one object of interest, one camera, and one main light source all directed toward the center of the object. The light is a fixed strength spotlight located at a fixed distance from the object. The material of each object is semi-metallic, which is common for the classes under consideration.  
The 3D collection consists of 100 shapes of 12 object classes (aeroplane, bench, bicycle, boat, bottle, bus, car, chair, dining table, motorbike, train, truck) from Pascal-3D \cite{pascal3D} and ShapeNet \cite{shapenet}.
At each iteration, one shape from the intended class is randomly picked and  placed in the middle of the scene. Then, the Blender rendered image is passed to YOLOV3 for detection. Please refer to \figLabel \ref{fig:qualitive_dataset_yolo} for some sample images of the object detection dataset. 

\begin{figure}[t]
\includegraphics[width=\columnwidth]{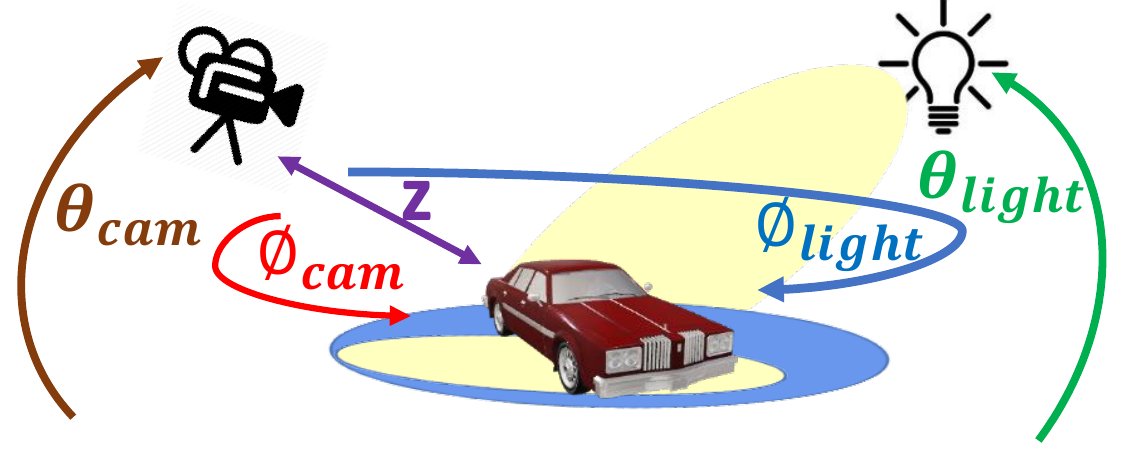}
\caption{\small \textbf{Visualization view point parameters.} These parameters are used in the first analysis experiment. The object class and RGB colors are fixed.}
\label{fig:analysis}
\end{figure}

\mysection{Environment parameters}
We use eight parameters that have shown to affect detection performance and frequently occur in real setups (refer to \figLabel \ref{fig:analysis}). The object is centered in the virtual scene, and the camera circles around the object keeping the object in the center of the rendered image. 
The parameters were normalized to $[-1,1]^{8}$ before using them for learning and testing.

\subsection{Self-Driving} \label{selfdrive}
There is a lot of recent work in autonomous driving especially in the fields of robotics and computer vision \cite{Franke2017,Codevilla2018}. In general, complete driving systems are very complex and difficult to analyze or simulate. By learning the underlying distribution of failure cases, our work provides a safe way to analyze the robustness of such a complete system. While our analysis is done in simulation only, we would like to highlight that sim-to-real transfer is a very active research field nowadays
\cite{Sadeghi2017,Tobin2017}.

\mysection{Agent}
We use an autonomous driving agent (based on CIL \cite{Codevilla2018}), which was trained on the environment $\mathbf{E}_{\mu}$ with default parameters. The driving-policy was trained end-to-end to predict car controls given an input image and is conditioned on high-level commands (\eg \emph{turn right at the next intersection}) in order to facilitate autonomous navigation.

\begin{figure*}
\centering
\includegraphics[width=\textwidth]{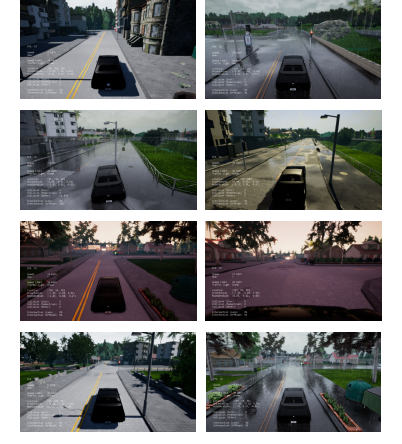}
\caption{\textbf{Environment for Self-driving} - Some samples of the CARLA \cite{carla} simulator environment.}
\label{fig:carla_env}
\end{figure*}

\mysection{Environment}
We use the recent CARLA driving simulator \cite{carla}, the most realistic open-source urban driving simulator currently available. We consider the three common tasks of driving in a straight line, completing one turn, and navigating between two random points. The score is measured as the average success of five pairs of start and end positions. \figLabel \ref{fig:carla_env} shows some images of the CARLA \cite{carla} simulation environment used for the self-driving car experiments.

\begin{figure*}
\centering
\includegraphics[width=0.5\textwidth]{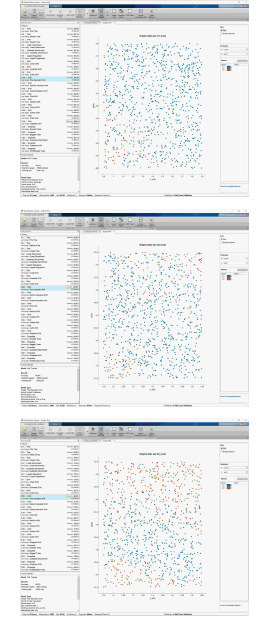}
\caption{\textbf{Training Data for Self-driving} - Visualization of training data distribution for 2 parameters (camera height, camera pitch angles) .}
\label{fig:carla_data}
\end{figure*}

\mysection{Environment parameters}
Since experiments are time-consuming, we restrict ourselves to three parameters, two of which pertain to the mounted camera viewpoint and the third  controls the appearance of the environment by changing the weather setting (\eg 'clear noon', 'clear sunset', 'cloudy after rain', etc.). As such, we construct an environment by randomly perturbing the position and rotation of the default camera along the z-axis and around the pitch axis respectively, and by picking one of the weather conditions. Intuitively, this helps measure the robustness of the driving policy to the camera position (\eg deploying the same policy in a different vehicle) and to environmental conditions. \figLabel \ref{fig:carla_data} visualizes the distribution of the training data.

\subsection{UAV Racing} \label{racingtrack}
In recent years, UAV (unmanned aerial vehicle) racing has emerged as a new sport where pilots compete in navigating small UAVs through race courses at high speeds. Since this is a very interesting research problem, it has also been picked up by the robotics and vision communities \cite{pmlr-v87-kaufmann18a}. 

\mysection{Agent}
We use a fixed agent to autonomously fly through each course and measure its success as percentage of gates passed \cite{TeachingUAVstoRace}. If the next gate was not reached within 10 seconds, we reset the agent at the last gate. We also record the time needed to complete the course. The agent uses a perception network that produces waypoints from image input and a PID controller to produce low-level controls. 

\begin{figure*}
\centering
\includegraphics[width=\textwidth]{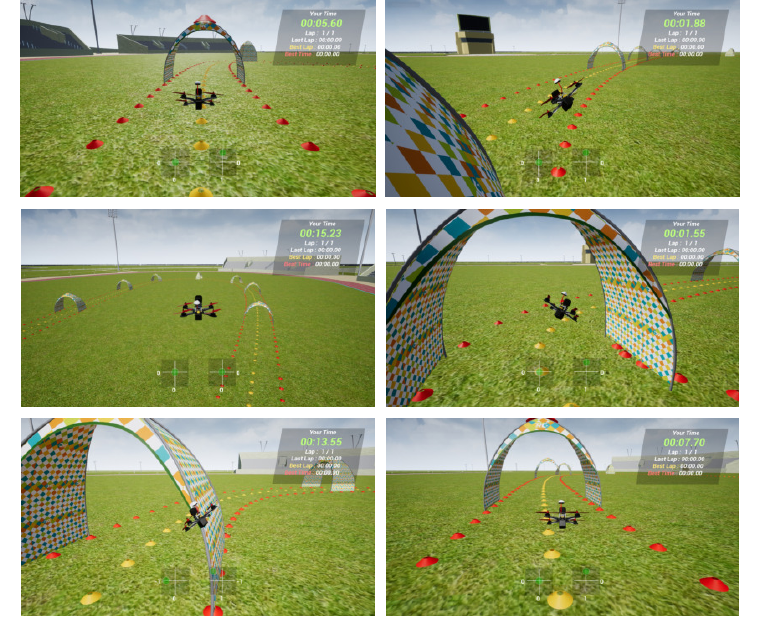}
\caption{\textbf{Environment for UAV Racing} - Some samples of the Sim4CV \cite{sim4cv} simulator environment.}
\label{fig:sim4cv_env}
\end{figure*}

\mysection{Environment}
We use the general-purpose simulator for computer vision applications, Sim4CV \cite{sim4cv}. Sim4CV is not only versatile but also photo-realistic and provides a suitable environment for UAV racing. \figLabel \ref{fig:sim4cv_env} shows some images of the Sim4CV simulator used for the UAV racing application.

\begin{figure*}
\centering
\includegraphics[width=\textwidth]{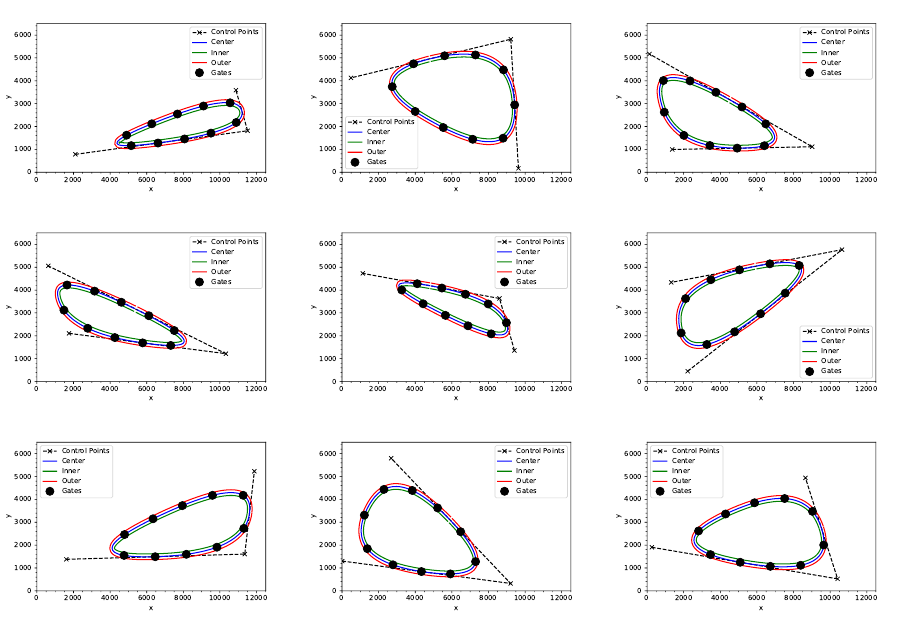}
\captionof{figure}{\textbf{Training Data for 3-Anchors UAV Racing} - Some sample tracks from the dataset with 3 control points.}
\label{fig:qualitive_results_uav_data_3}
\end{figure*}

\begin{figure*}
\centering
\includegraphics[width=\textwidth]{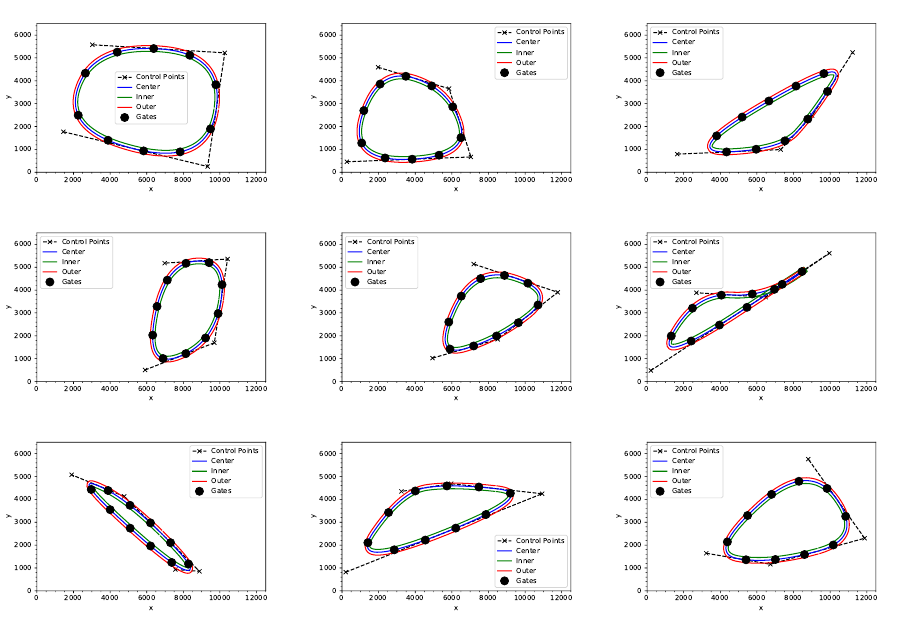}
\captionof{figure}{\textbf{Training Data for 4-Anchors UAV Racing} - Some sample tracks from the dataset with 4 control points.}
\label{fig:qualitive_results_uav_data_4}
\end{figure*}

\begin{figure*}
\centering
\includegraphics[width=\textwidth]{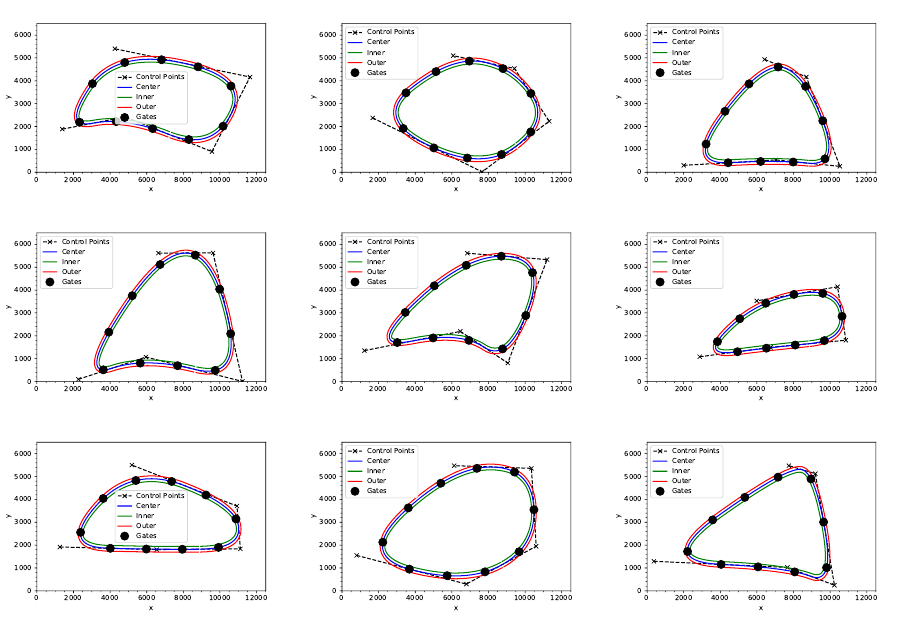}
\captionof{figure}{\textbf{Training Data for 5-Anchors UAV Racing} - Some sample tracks from the dataset with 5 control points.}
\label{fig:qualitive_results_uav_data_5}
\end{figure*}

\mysection{Environment parameters}
We change the geometry of the race course environment. We define three different race track templates with 3-5 2D anchor points, respectively. These points describe a second order B-spline and are perturbed to generate various race tracks populated by uniformly spaced gates. Figures \ref{fig:qualitive_results_uav_data_3}, \ref{fig:qualitive_results_uav_data_4} and \ref{fig:qualitive_results_uav_data_5} show some samples from the UAV datasets.

\clearpage
\section{Analysis} \label{sup:analysis}
\subsection{Diagnosis} \label{sup:diagnosis}
To identify weaknesses and cases that results in systematic failure for the YOLOv3 detector, we fix some semantic parameters and attack the others. We conduct two different case studies. The first one involves the camera view-point and light direction; these have the largest impact in determining the pixel values of the final image. The second case study examines occlusion, which is one of the most common causes of detection failure. In both cases, we focus on two classes that are relevant in the autonomous driving application: cars and motorbikes. Since we have several 3D models for each class, we include the effect of model shapes in the analysis. 
We use roughly sampled models with homogeneous texturing (obtained from the training in \secLabel{\ref{sec:application-detection}}) as well as detailed fully-textured models. This also has the nice side-effect to show how well our insights transfer to more realistic renderings and ultimately the real world. 
In total, we consider five different scenarios per case. \textbf{Scenario 1:} simple car, \textbf{Scenario 2:} detailed car, \textbf{Scenario 3:} simple motorbike, \textbf{Scenario 4:} detailed motorbike 1, \textbf{Scenario 5:}  detailed motorbike 2. We use these scenarios for both cases.        

\begin{figure}[!htb]
\includegraphics[width=\columnwidth]{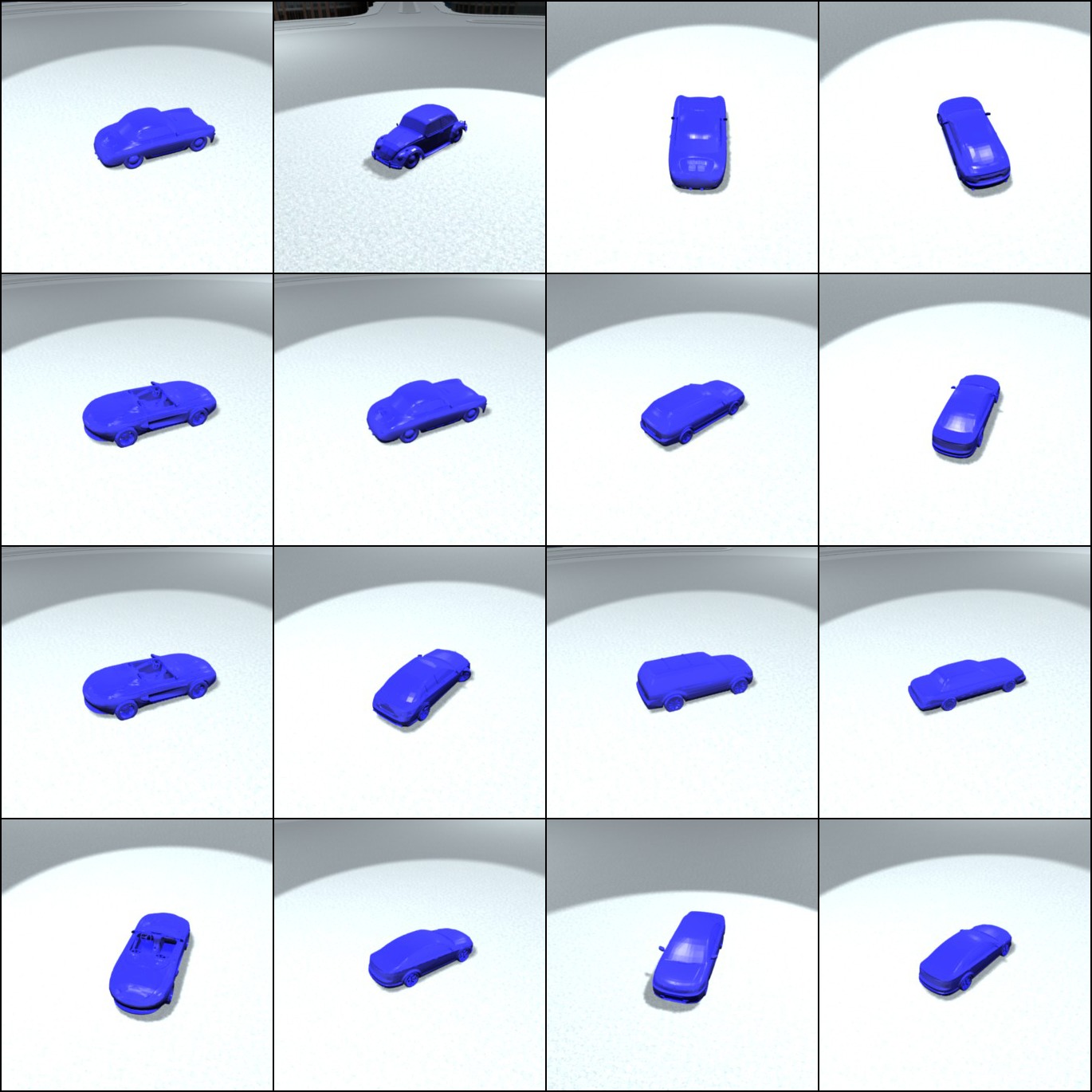}
\caption{\small \textbf{Case 1, Scenario 1, Qualitative Examples: }generated by BBGAN}
\label{fig:scenario1}
\end{figure}

\begin{figure}[!htb]
\includegraphics[width=\columnwidth]{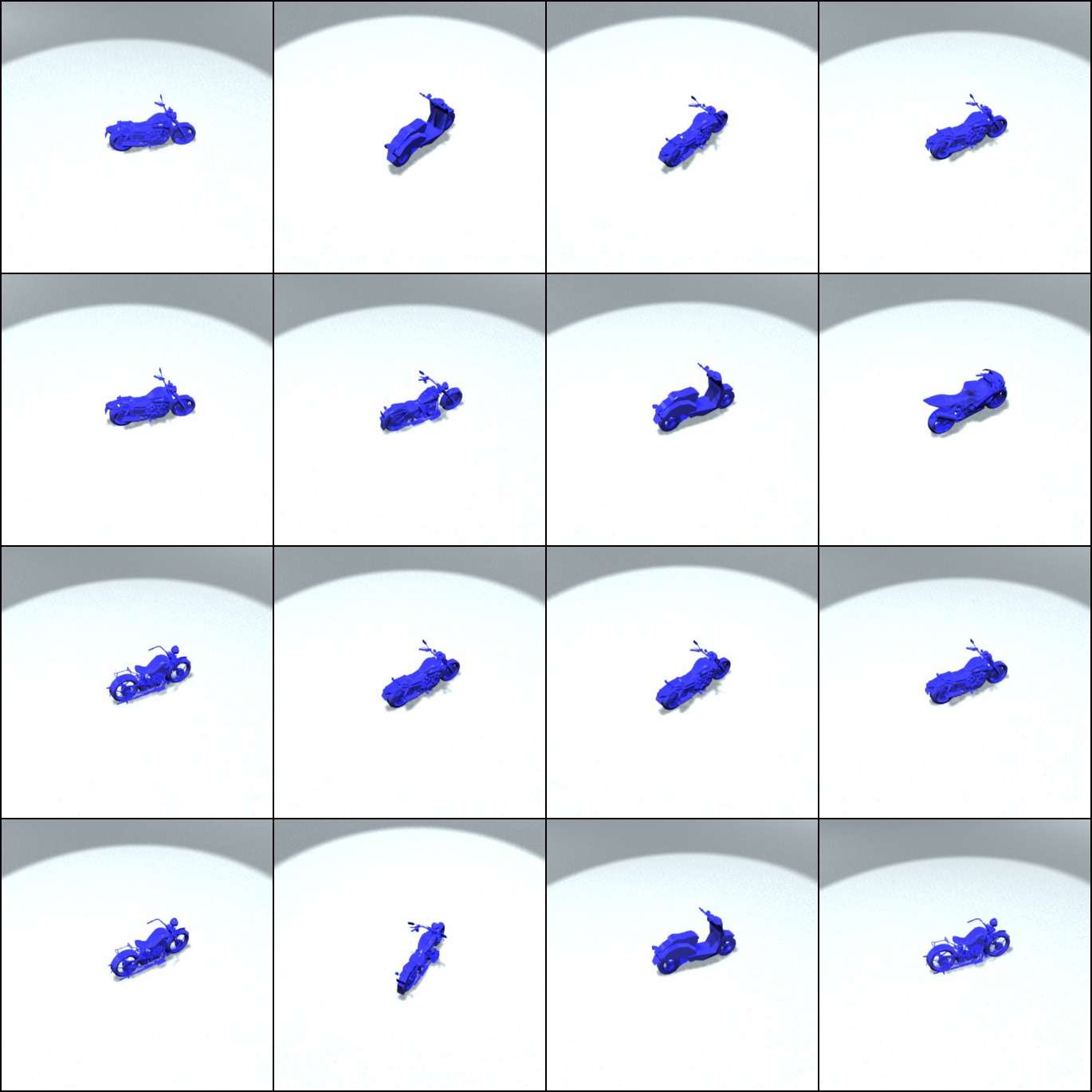}
\caption{\small \textbf{Case 1, Scenario 2, Qualitative Examples: }generated by BBGAN}
\label{fig:scenario2}
\end{figure}

\begin{figure}[!htb]
\includegraphics[width=\columnwidth]{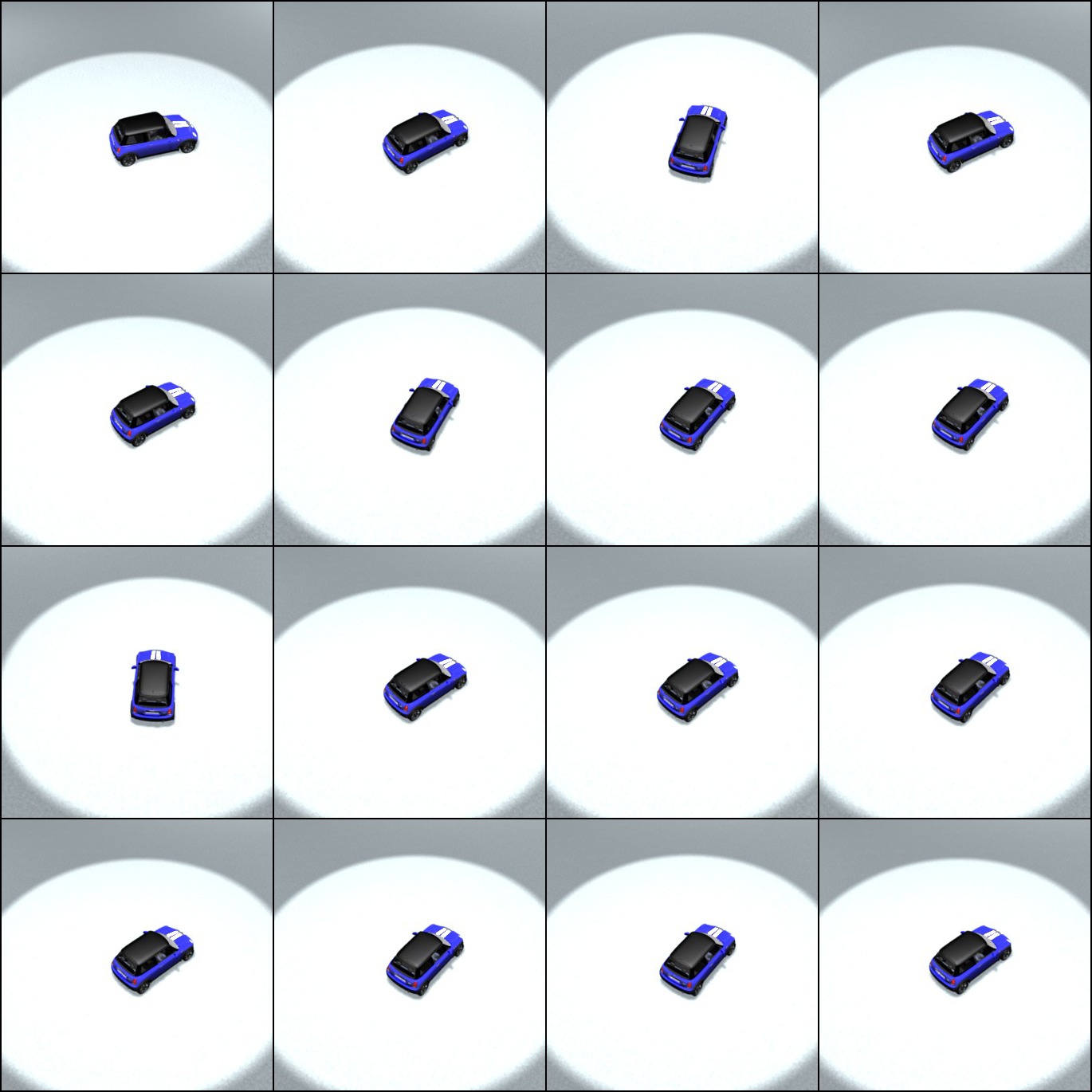}
\caption{\small \textbf{Case 1, Scenario 3, Qualitative Examples: }generated by BBGAN}
\label{fig:scenario3}
\end{figure}

\begin{figure}[!htb]
\includegraphics[width=\columnwidth]{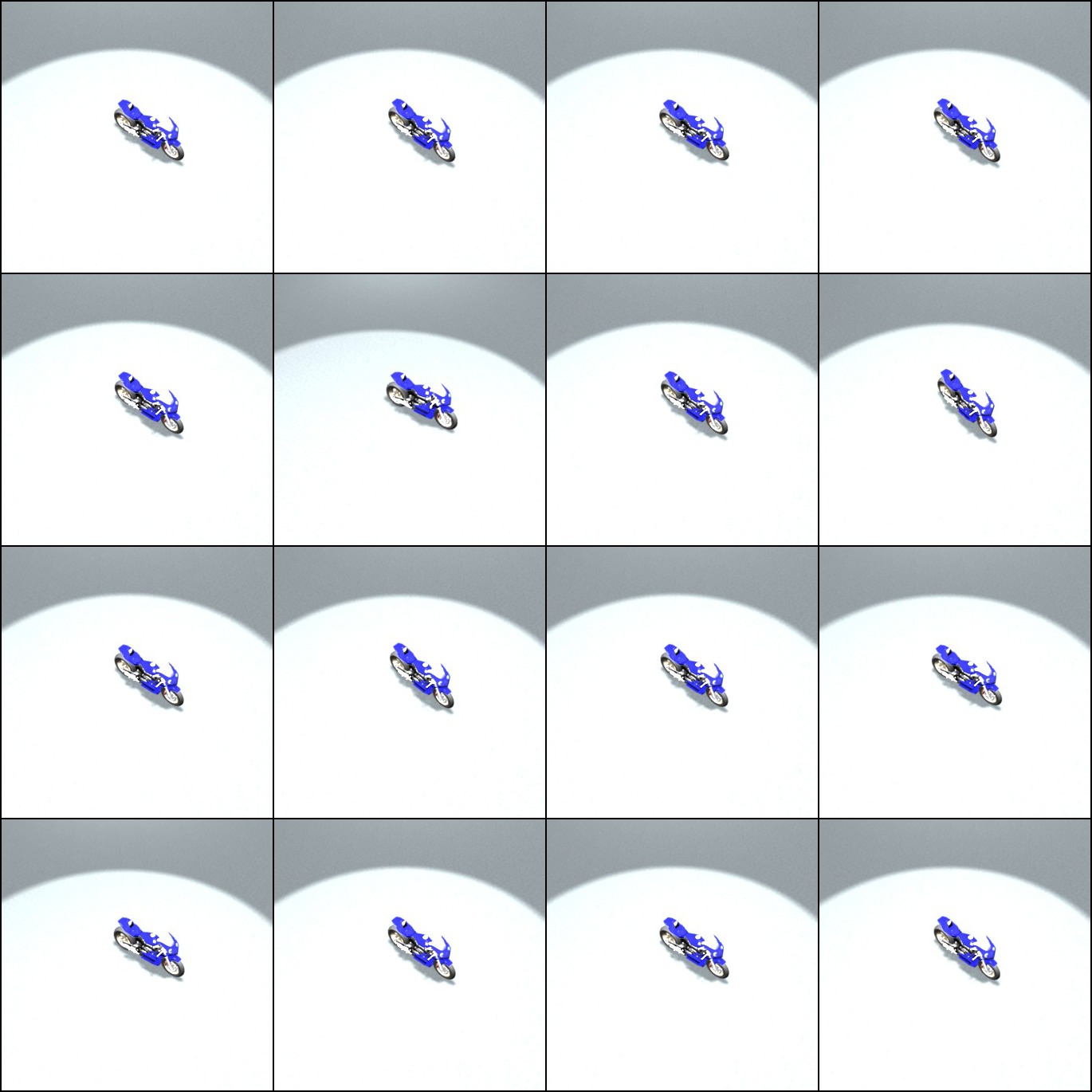}
\caption{\small \textbf{Case 1, Scenario 4, Qualitative Examples: }generated by BBGAN}
\label{fig:scenario4}
\end{figure}

\begin{figure}[!htb]
\includegraphics[width=\columnwidth]{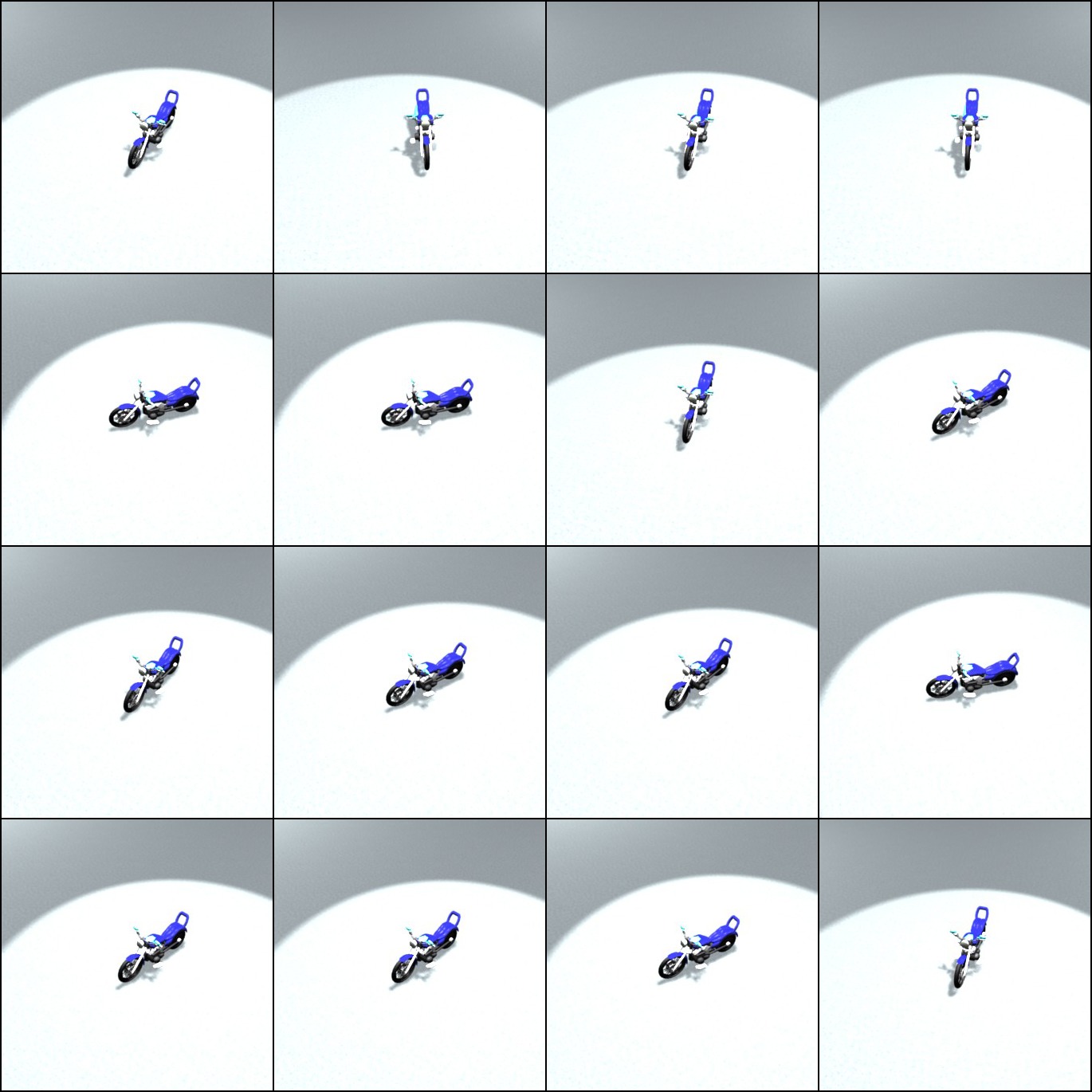}
\caption{\small \textbf{Case 1, Scenario 5, Qualitative Examples: }generated by BBGAN}
\label{fig:scenario5}
\end{figure}

\vspace{3pt}\mysection{Case 1: View point}
We restrict the number of parameters to 4 ($\phi_{\text{cam}},\theta_{\text{cam}},\phi_{\text{light}},\phi_{\text{light}}$), and fix the object class and RGB colors (pure blue). Figures \ref{fig:scenario1}, \ref{fig:scenario2}, \ref{fig:scenario3}, \ref{fig:scenario4} and \ref{fig:scenario5} show qualitative results for samples generated by learning a BBGAN on each scenario in the view-point case. Figures \ref{fig:analysis-car-sup} and \ref{fig:analysis-motorbike-sup} visualize the learned distribution in scenario 3 and scenario 5 and some examples of transferability to real world.

\begin{figure}[!htb]
\includegraphics[width=\columnwidth]{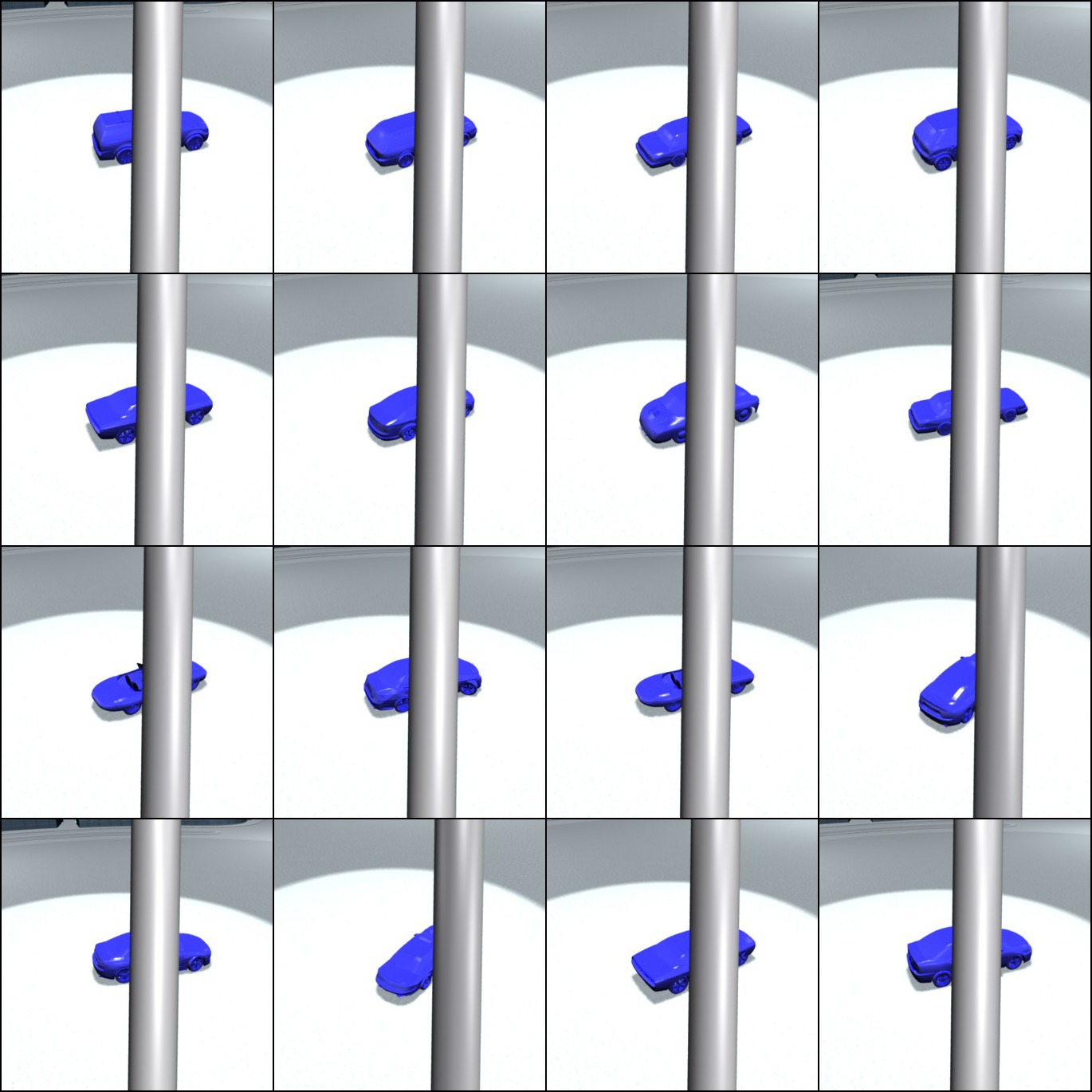}
\caption{\small \textbf{Case 2, Scenario 1, Qualitative Examples: }generated by BBGAN}
\label{fig:scenario6}
\end{figure}

\begin{figure}[!htb]
\includegraphics[width=\columnwidth]{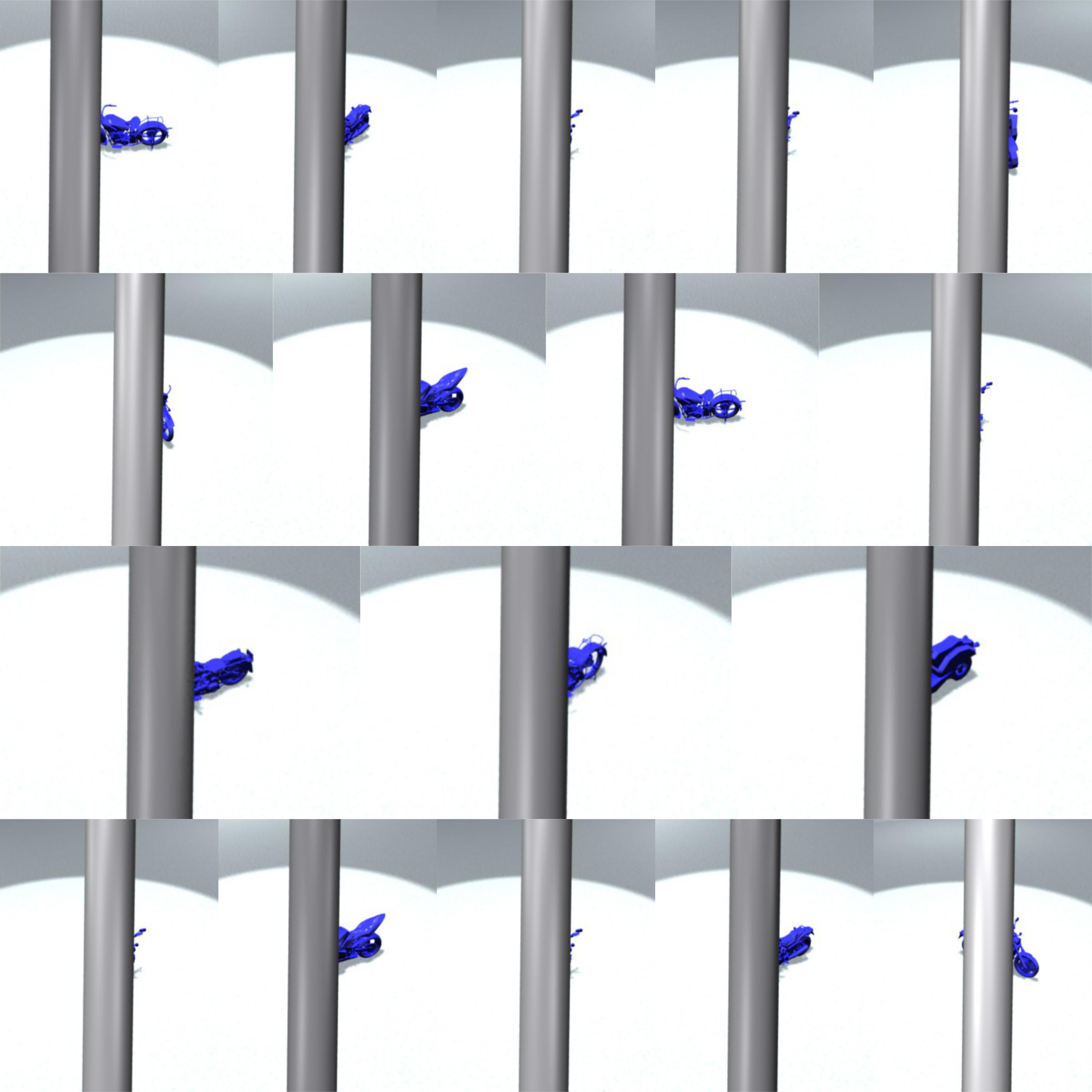}
\caption{\small \textbf{Case 2, Scenario 2, Qualitative Examples: }generated by BBGAN}
\label{fig:scenario7}
\end{figure}

\begin{figure}[!htb]
\includegraphics[width=\columnwidth]{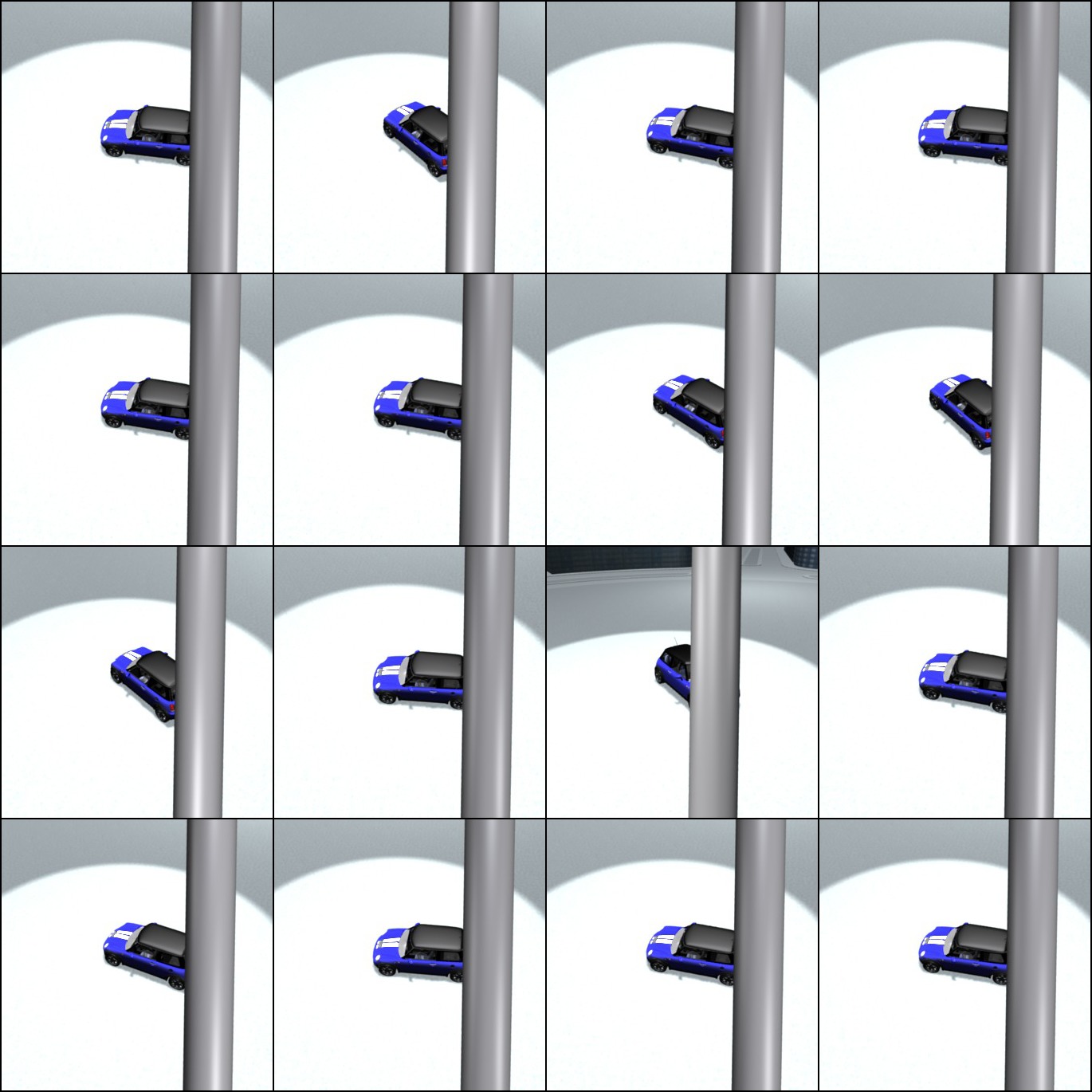}
\caption{\small \textbf{Case 2, Scenario 3, Qualitative Examples: }generated by BBGAN}
\label{fig:scenario8}
\end{figure}

\begin{figure}[!htb]
\includegraphics[width=\columnwidth]{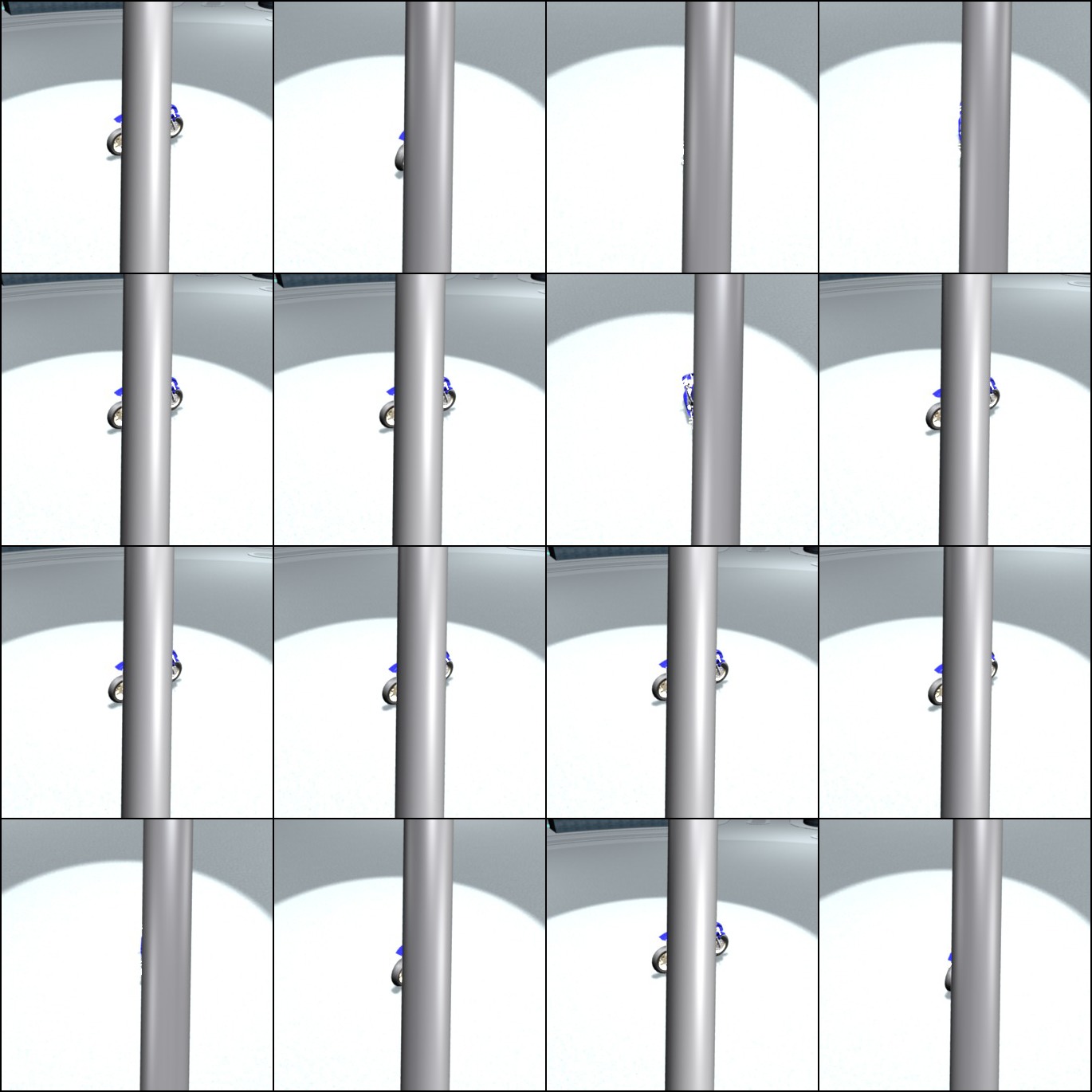}
\caption{\small \textbf{Case 2, Scenario 4, Qualitative Examples: }generated by BBGAN}
\label{fig:scenario9}
\end{figure}

\begin{figure}[!htb]
\includegraphics[width=\columnwidth]{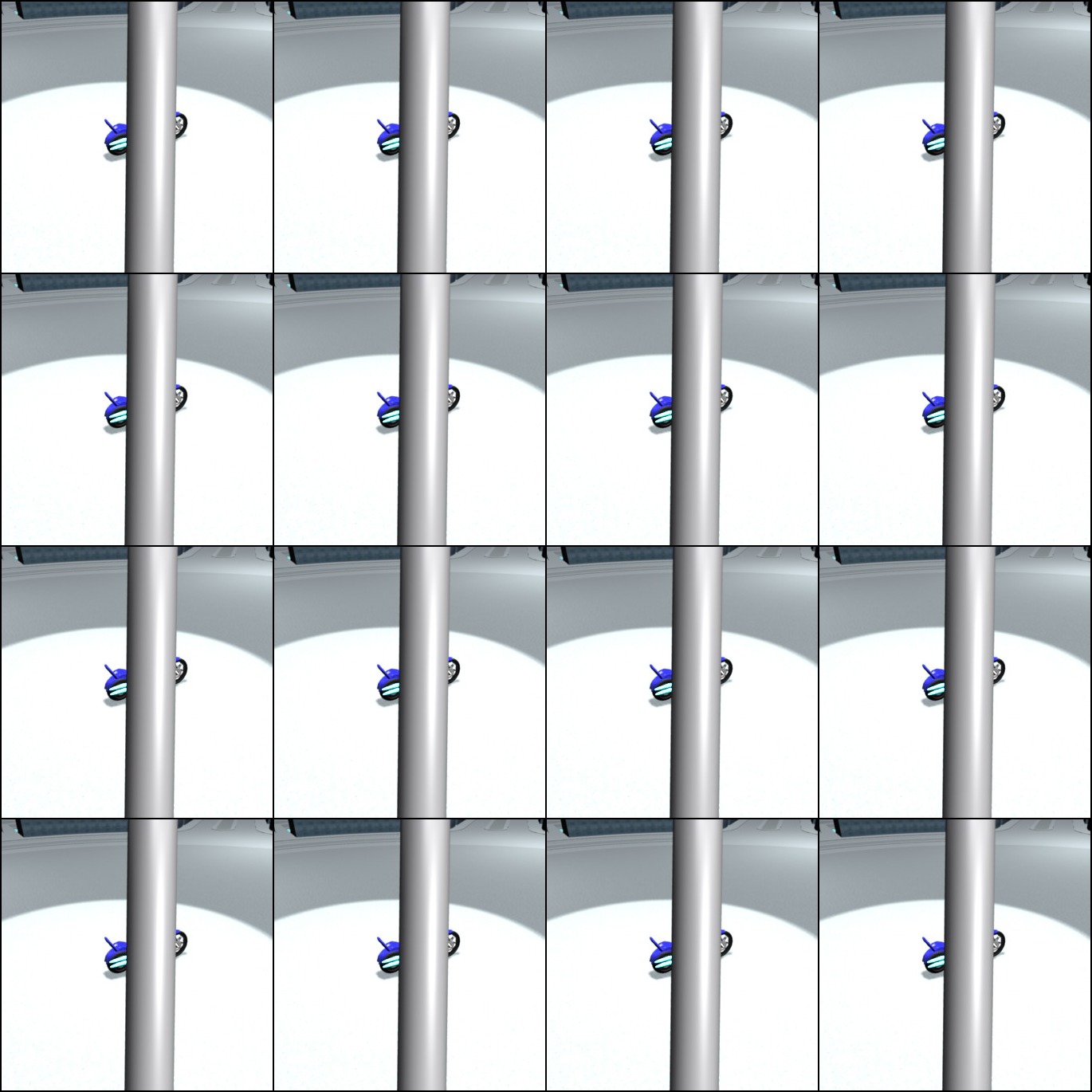}
\caption{\small \textbf{Case 2, Scenario 5, Qualitative Examples: } generated by BBGAN}
\label{fig:scenario10}
\end{figure}

\vspace{3pt}\mysection{Case 2: Occlusion}
Since occlusion plays an important role in object misdetection, we introduce an occlusion experiment. Here, we investigate how occlusion (\eg by a pole) can result in failure of a detector (\eg from which view point). Therefore, we include the camera viewpoint angles ($\phi_{\text{cam}},\theta_{\text{cam}}$) and introduce a third parameter to control horizontal shift of a pole that covers 15\% of the rendered image and moves from one end to another. The pole keeps a fixed distance to the camera and is placed between the camera and the object. Figures \ref{fig:scenario6}, \ref{fig:scenario7}, \ref{fig:scenario8}, \ref{fig:scenario9} and \ref{fig:scenario10} show qualitative results for samples generated by learning a BBGAN on each scenario in the occlusion case. 

\subsection{Transferability}
 
\mysection{Across Shape Variations}\\
We use the same scenarios as above to construct transferability experiments. The goal is to validate generalization capabilities of the learned fooling distribution from one scenario to another. Also, it shows what role the model shape plays with regard to the strength of the learned attacks. Tables \ref{case1-table} and \ref{case2-table} show the transferability of the adversarial attacks for Case 1 and Case 2. We see that most attacks transfer to new scenarios that are similar, indicating the generalization of the learned fooling distribution. However, the attacks that were learned on more detailed CAD models transfer better to generic less detailed models (\eg PASCAL3D\cite{pascal3D} and ShapeNet\cite{shapenet} models).

To validate generalization capabilities of the learned fooling distribution, we learn this distribution from samples taken from one setup and then test it on another. 
Table \ref{case1-table} shows that adversarial distributions learned from detailed textured models transfer better (\ie maintain similar AFR after transferring to the new setup) than those learned from rough ones from Pascal3D \cite{pascal3D} and ModelNet \cite{modelnet}. This observation is consistent with that of \cite{strike}. 

\begin{table}[h]
\small
\tabcolsep=0.09cm
\centering
 \begin{tabular}{||c  c c|c  c c ||} 
 \hline
  \multicolumn{3}{||c|}{original attacks} & \multicolumn{3}{|c||}{transferred attacks}\\ 
   \hline  \hline
scenario \#	& BBGAN 	& random	& to scenario \#	& BBGAN  &     random  \\
\hline
1	& 96.4\%	& 70.4\%	& 2		& 0.0\%	& 26.0\% 		\\
2	& 88.8\%	& 26.0\%	& 1		& 90.4\%	& 70.4\%	\\
3	& 92.0\%	& 53.2\%	& 4		& 0.8\%	& 10.4\%		\\
3	& 92.0\%	& 53.2\%	& 5		& 3.2\%	& 13.2\%	\\
4	& 90.8\%	& 10.4\%	& 3		& 95.2\%	& 53.2\%	\\
5	& 16.8\%	& 13.2\%	& 3		& 64.8\%	& 53.2\%	\\
[1ex] 
 \hline 
 \end{tabular}
\caption{\textbf{Case 1: view-point attack transferability:}  Attack Fooling Rate for sampled attacks on each scenario and transferred attacks from one scenario to another. Random attacks for each scenario are provided for reference.}
\label{case1-table}
\end{table}

\begin{table}[h]
\small
\tabcolsep=0.09cm
\centering
 \begin{tabular}{||c  c c|c  c c ||} 
 \hline
  \multicolumn{3}{||c|}{original attacks} & \multicolumn{3}{|c||}{transferred attacks}\\ 
   \hline  \hline
scenario \#	& BBGAN 	& random	& to scenario \#	& BBGAN  &     random  \\
\hline
1	& 96.8\%	& 58.0\%	& 2		& 36.0\%	& 32.4\%	\\
2	& 94.8\%	& 32.4\%	& 1		& 77.2\%	& 58.0\%	\\
3	& 95.6\%	& 52.0\%	& 4		& 90.8\%	& 37.2\%	\\
3	& 95.6\%	& 52.0\%	& 5		& 94.0\%	& 39.6\%	\\
4	& 99.2\%	& 37.2\%	& 3		& 100.0\%	& 52.0\%	\\
5	& 100.0\%	& 39.6\%	& 3		& 100.0\%	& 52.0\%	\\
[1ex] 
 \hline 
 \end{tabular}
\caption{\textbf{Case 2: occlusion attack transferability.} Attack Fooling Rate for sampled attacks on each scenarios and also for transferred attacks from one scenario to another. Random attacks for each scenario are put for reference}
\label{case2-table}
\end{table}

\begin{figure}
\includegraphics[width=\columnwidth]{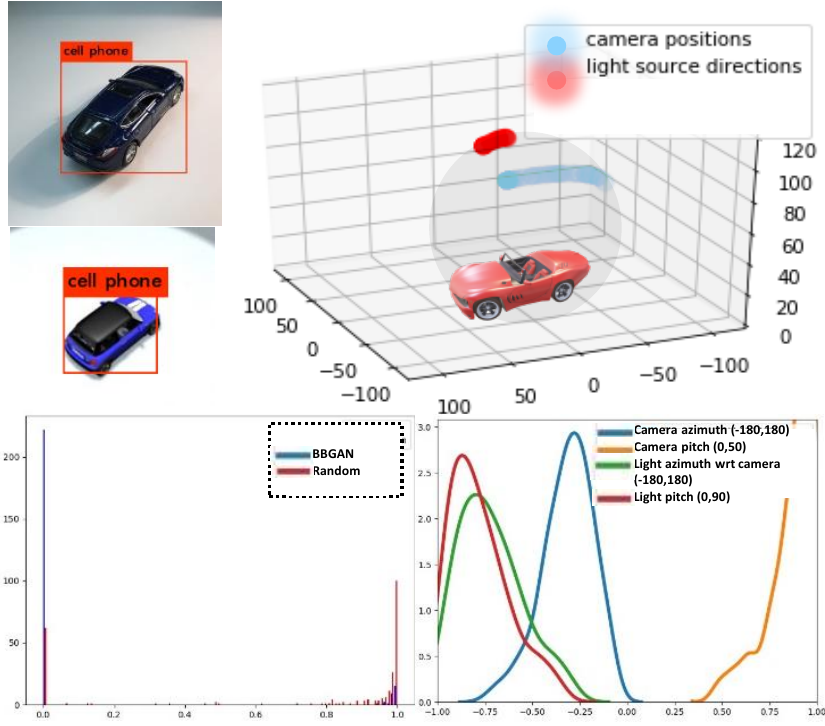}
\caption{\small \textbf{Analysis: Visualization of the Fooling Distribution}. We fix the object to be a car and fix the distance to the camera and train a BBGAN to learn the fooling camera and light source angles to fool the YOLOV3 detector. \textit{Top}: on the right we plot the camera positions and light source directions of 250 sampled parameters in a 3D sphere around the object. On the left we show how taking real photos from the same rendered angles of some toy car confuses the YOLOV3 detector as the rendered image. \textit{Bottom}: on the right  we visualize the distribution of parameters normalized from (-1,1), while on the left we visualize the histogram of scores (0 to 1) of the learned parameters distribution vs random distribution. }
\label{fig:analysis-car-sup}
\end{figure}

\begin{figure}
\includegraphics[width=\columnwidth]{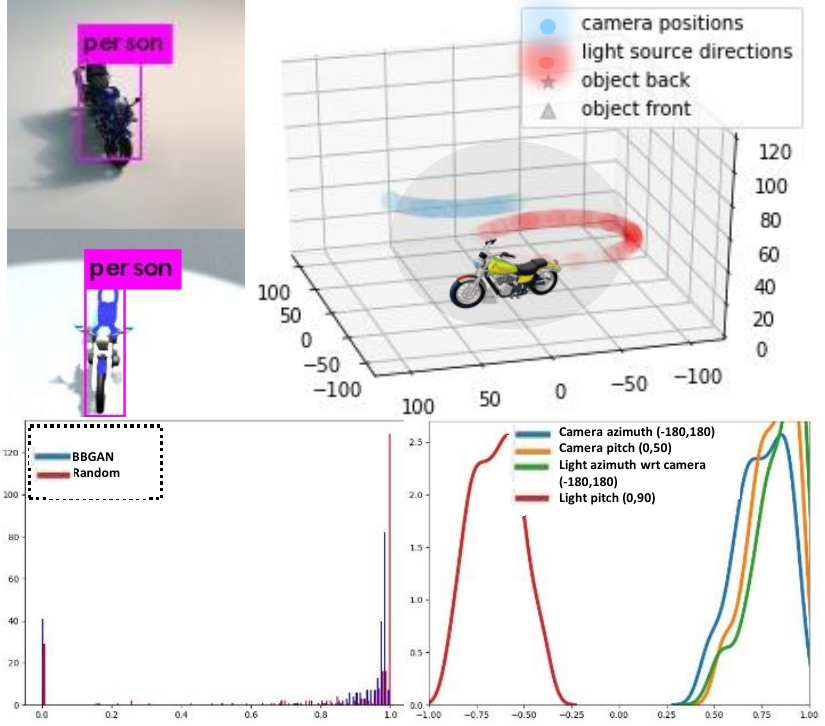}
\caption{\small \textbf{Analysis: Visualization of the Fooling Distribution}. We fix object to be a motorbike and fix the distance to the camera and train a BBGAN to learn the fooling camera and light source angles to fool the YOLOV3 detector. \textit{Top}: on the right we plot the camera positions and light source directions of 250 sampled parameters in a 3D sphere around the object. On the left we show how taking real photos from the same rendered angles of some toy motorbike confuses the YOLOV3 detector as the rendered image. \textit{Bottom}: on the right  we visualize the distribution of parameters normalized from (-1,1), while on the left we visualize the histogram of scores (0 to 1) of the learned parameters distribution vs random distribution. }
\label{fig:analysis-motorbike-sup}
\end{figure}

\mysection{Virtual to Real World}\\
To demonstrate the transferability of the fooling parameter distribution to the real world, we photograph a toy motorbike, similar to the 3D model we are attacking. We use a mobile phone camera and an office spotlight to replace the light source in the virtual environment. The photos are taken under different camera views and lighting directions (uniform sampling). We also take photos based on samples from the distribution learned by the BBGAN. We apply the YOLOv3 detector on these images and observe the confidence score for the `motorbike' class of interest. 
On the samples generated from the BBGAN distribution, the attack fooling rate is 21\% compared to only 4.3\% when picking a random viewpoint. In Figures \ref{fig:analysis-car-sup} and \ref{fig:analysis-motorbike-sup}, we visualize the fooling distribution generated by our BBGAN and provide some corresponding real-world images.

\mysection{Visualization}
In Figures \ref{fig:analysis-car-sup} and \ref{fig:analysis-motorbike-sup}, we visualize of the fooling distribution generated by our BBGAN in the two previous experiments (\secLabel{\ref{sup:diagnosis}}). We also include some real-world images captured according to the parameters generated by the BBGAN.

\section{Insights Gained by BBGAN Experiments}
\subsection{Object Detection with YOLOV3}
In our YOLOV3 experiments, we consistently found that for most objects top rear or top front views of the object are fooling the YOLOV3 detector. Furthermore, the light angle which will result in highest reflection off the surface of the object also results in higher fooling rates for the detector. The color of the object does not play a big role in fooling the detector, but usually colors that are closer to the background color tend to be preferred by the BBGAN samples (as shown in the qualitative examples). From the analysis in \secLabel {\ref{sup:analysis}} of transferability of these attacks, we note that attacks on more detailed CAD shapes and models transfer better to less detailed shapes, but the opposite is not true. 

\subsection{Self-driving cars}
In our experiments we found that weather is the least important parameter for determining success. This is probably due to the fact that the driving policy was trained on multiple weather conditions. This allows for some generalization and robustness to changing weather conditions. However, the driving policy was trained with a fixed camera. We observe, that the driving policy is very sensitive to slight perturbations of the camera pose (height and pitch).

\subsection{UAV Autonomous Navigation}
We observe that the UAV fails if the track has very sharp turns. This makes intuitive sense and the results that were produced by our BBGAN consistently produce such tracks. For the tracks that are only parameterized by three control points it is difficult to achieve sharp turns. However, our BBGAN is still able to make the UAV agent fail by placing the racing gates very close to each other, thereby increasing the probability of hitting them. 

\end{document}